\documentclass[11pt]{article}
\usepackage{arxiv}

\usepackage[utf8]{inputenc}
\usepackage[T1]{fontenc}
\usepackage{lmodern}

\usepackage{amsmath,amssymb,amsfonts}
\usepackage{booktabs}
\usepackage{multirow}
\usepackage[table]{xcolor}
\usepackage{subcaption}
\usepackage{graphicx}
\usepackage{float}
\usepackage{url}
\usepackage[normalem]{ulem}
\usepackage{microtype}
\usepackage{enumitem}
\usepackage{natbib}
\usepackage[pdftex,colorlinks]{hyperref}

\title{A Unifying Human-Centered AI Fairness Framework}

\author{
  Munshi Mahbubur Rahman\thanks{Corresponding author.} \\
  Department of Information Systems \\
  University of Maryland, Baltimore County \\
  Baltimore, MD, USA \\
  \texttt{mrahman4@umbc.edu}
  \and
  Shimei Pan \\
  Department of Information Systems \\
  University of Maryland, Baltimore County \\
  Baltimore, MD, USA \\
  \texttt{shimei@umbc.edu}
  \and
  James R.~Foulds \\
  Department of Information Systems \\
  University of Maryland, Baltimore County \\
  Baltimore, MD, USA \\
  \texttt{jfoulds@umbc.edu}
}

\begin{document}

\maketitle

\begin{abstract}
The increasing use of Artificial Intelligence (AI) in critical societal domains has amplified concerns about fairness, particularly regarding unequal treatment across sensitive attributes such as race, gender, and socioeconomic status. While there has been substantial work on ensuring AI fairness, navigating trade-offs between competing notions of fairness as well as predictive accuracy remains challenging, which is a barrier to the practical deployment of fair AI systems. To address this, we introduce a unifying human-centered fairness framework that systematically covers eight distinct fairness metrics, formed by combining individual vs.\ group fairness, infra-marginal vs.\ intersectional assumptions, and outcome-based vs.\ equality-of-opportunity (EOO) options, thereby allowing stakeholders to align fairness interventions with their value systems and contextual considerations. The framework uses a consistent and easy to understand formulation for all metrics to reduce the learning curve for non-expert stakeholders. Rather than privileging a single fairness notion, our framework enables stakeholders to assign weights across multiple fairness objectives, reflecting their priorities and values, and enabling multi-stakeholder compromises. We apply this approach to four real-world datasets—the UCI Adult census dataset for income prediction, the COMPAS dataset for criminal recidivism in the justice system, the German Credit dataset for credit risk assessment in financial services, and the MEPS dataset for healthcare utilization—and demonstrate how adjusting weights reveals nuanced trade-offs between different fairness metrics. Finally, through two stakeholder-grounded case studies in judicial decision-making and healthcare, we show how the framework can inform practical and value-sensitive deployment of fair AI systems.
\end{abstract}

\section{Introduction}
The widespread deployment of Artificial Intelligence (AI) systems in high-stakes domains—such as healthcare, criminal justice, and financial services—has raised significant concerns about fairness and bias \cite{angwin2016machine,obermeyer2019dissecting,barocas2020hidden}. Studies have shown that these systems can inherit, perpetuate, or even exacerbate historical and systemic biases present in their training data or decision-making pipelines, leading to disparate outcomes across demographic groups. For instance, risk assessment tools used in the criminal justice system have been shown to assign higher recidivism risk scores to Black defendants compared to White defendants with similar profiles \cite{angwin2016machine}. In healthcare, widely-used predictive models have demonstrated racial bias in prioritizing patients for care \cite{obermeyer2019dissecting}.

Although there is increasing research focused on strategies to mitigate bias in AI, these approaches have yet to become widely adopted in the practical implementation of AI systems in various industries, governmental bodies, and the public sector~\cite{ainow2023landscape}. One significant reason for this limited adoption is the intricate nature of fairness itself. Achieving fairness in AI systems requires reconciling conflicting technical definitions of fairness and the underlying societal values they represent. This challenge arises from inherent conflicts among fairness definitions and the trade-offs between fairness and predictive performance, creating a dilemma for developers and organizations striving to design equitable systems. There is also the concern that an intense focus on fairness may adversely affect the accuracy of AI systems, creating a perceived trade-off between fairness and predictive performance \cite{hardt2016equality}. In this work, we address these challenges through a unifying fairness optimization framework that enables stakeholders to navigate and balance conflicting fairness criteria.

The complexity of achieving fairness lies in balancing multiple, often incompatible, fairness metrics and addressing the societal assumptions underlying them~\cite{berk2021fairness}. Identifying fairness metrics that align with different societal values and assumptions is often the starting point for fairness interventions in AI systems~\cite{blodgett2020language}. Broadly, AI fairness definitions can be categorized along two dimensions: the granular level of protection and the approach to addressing disparities. Individual fairness emphasizes providing comparable outcomes for individuals with similar qualifications, whereas group fairness ensures equitable treatment across demographic groups \cite{dwork2012fairness}. Infra-marginal perspectives \cite{ayres2002outcome} often interpret existing disparities as reflections of legitimate group differences and therefore consider them acceptable, whereas intersectional approaches view these disparities as manifestations of systemic inequities that require intervention~\cite{foulds2020parity}.

Building on these foundations, this paper introduces a unifying human-centered fairness framework that incorporates a richer set of fairness metrics—spanning both prediction outcome-based and error-rate-based perspectives—and enables multiple stakeholders to assign weights to different fairness goals. Prediction-based metrics assess model outputs such as classification probabilities, while EOO-based metrics examine disparities in false positives. Unlike typical approaches that optimize a single fairness criterion in isolation, our framework enables principled balancing between competing fairness definitions and predictive performance. Each metric is expressed in a consistent mathematical formulation, clarifying their relationships and easing stakeholder understanding. The framework explicitly supports metrics from both individual and group fairness paradigms, as well as from infra-marginal and intersectional perspectives.

Our goal is to empower stakeholders to surface conflicts and navigate them by finding sensible compromises. We validate our framework through experiments on four benchmark datasets—the UCI Adult census dataset, COMPAS recidivism dataset, German Credit dataset, and MEPS healthcare dataset—and detailed case studies. A preliminary version appeared in \cite{rahman2024towards}; this journal article extends that work with EOO-based metrics, expanded experiments, and stakeholder-grounded case studies. Our main contributions are:
\begin{itemize}
    \item \textit{A unified, stakeholder-centered fairness framework} spanning individual vs. group and infra-marginal vs. intersectional perspectives, with outcome- and EOO-based metrics under a shared formulation.
    \item \textit{A multi-objective optimization strategy} allowing stakeholders to assign weights to fairness criteria and accuracy, enabling consensus solutions.
    \item Empirical validation across four datasets with case studies, demonstrating navigation of fairness–performance trade-offs and stakeholder-aligned decisions.
\end{itemize}

\section{Background}
Achieving fairness in AI necessitates addressing societal inequities and conflicting philosophical approaches to fairness. This section explores fairness from two key perspectives: societal theories of fairness and their implementation in AI systems.

\subsection{Social Theories of Fairness}
Our framework draws on two prominent fairness concepts that differ in their philosophical foundations and treatment of disparities: {\em intersectionality} \cite{truth1851aint,crenshaw2013demarginalizing,collins2022black} and {\em infra-marginality} \cite{ayres2002outcome,simoiu2017problem}.

{\em Intersectionality}, introduced by Kimberlé Crenshaw \cite{crenshaw2013demarginalizing} and expanded by Patricia Hill Collins \cite{collins2022black}, examines how overlapping systems of oppression create compounded disadvantages for individuals at the intersections of demographic groups, such as race, gender, and socioeconomic status. This framework emphasizes systemic inequities embedded within societal structures, leading to enduring disadvantages for marginalized groups. For instance, Collins and others \cite{collins2022black,collective1977black} have illustrated how intersecting factors like race, nationality, sexual orientation, disability, and social class interact to produce layered and unique forms of oppression. This politically progressive perspective identifies systemic inequities as the root causes of disparities and advocates for affirmative interventions to mitigate them. In the context of AI, intersectionality has been pivotal in exposing biases, such as how facial recognition systems often underperform for marginalized demographics~\cite{buolamwini2018gender}, and differential fairness has been proposed as an intersectional definition of AI fairness, focusing on parity at the intersection of protected subgroups~\cite{foulds2020intersectional,islam2023differential}.

In contrast, {\em infra-marginality} \cite{simoiu2017problem} assumes that differences in risk or ability distributions across demographic groups reflect legitimate individual-level variations stemming from effort, talent, or choice.\footnote{The term ``infra-marginality'' originally referred to a technical point, namely  that different populations' differing properties would naturally lead to disparities in overall ``risk,'' or supposed deservedness of an outcome~\cite{ayres2002outcome,simoiu2017problem}. Since ~\cite{ayres2002outcome,simoiu2017problem} tacitly assume these disparities are legitimate and fair, the concept thus takes on philosophical and political dimensions. Following Foulds and Pan~\cite{foulds2020parity}, we use the term in reference to the politically conservative values and assumptions that underpin it, i.e., an emphasis on meritocracy at the expense of parity.} The concept of \textit{infra-marginality} was introduced by~\cite{ayres2002outcome}, and was studied further by ~\cite{simoiu2017problem}. This perspective, rooted in conservative public policy, operates on the premise of a meritocratic societal baseline. Interventions aimed at equalizing outcomes are often resisted under this framework, which prioritizes predictive accuracy over equitable corrections. Critics, however, argue that infra-marginality neglects systemic barriers, such as racism, sexism, and economic inequality, that significantly contribute to observed disparities. For example, limited access to quality education, employment bias, and disproportionate law enforcement can severely impact outcomes for marginalized communities, underscoring the need to account for systemic inequalities \cite{collective1977black,crenshaw2013demarginalizing,davis2011prisons,hooks2014ain,wald2003defining}. Foulds and Pan~\cite{foulds2020parity} discussed the relative merits of parity-based fairness notions which run counter to infra-marginality arguments.

While much of the fairness discourse has focused on group-level disparities, philosophers also consider fairness at the individual level~\cite{dwork2012fairness}. Calsamiglia~\cite{calsamiglia2005decentralizing} states, ``Philosophers define equality of opportunity as the requirement that an \textit{individual’s} well being be
independent of his or her irrelevant characteristics.'' Calsamiglia further adds, ``The difference among philosophers is
mainly about which characteristics should be considered irrelevant.'' Dwork et al.~\cite{dwork2012fairness} proposed an individual fairness definition for AI systems.

\subsection{Fairness and Bias in AI}

Bias in AI systems often originates from the data on which these models are trained \cite{barocas2016big}. Historical and systemic inequities embedded in datasets perpetuate and even amplify existing disparities when used in algorithmic decision-making systems~\cite{zhao2017men}. Over the years, substantial research \cite{mehrabi2021survey,barocas-hardt-narayanan,hardt2016equality,vzliobaite2017measuring,dwork2012fairness,binns2018fairness} has been conducted to unpack these complexities and challenges associated with fairness in AI and machine learning systems. Scholars have explored a range of fairness metrics, methodologies, and applications aimed at addressing biases in such systems. These metrics broadly fall into two categories: \emph{individual} fairness and \emph{group} fairness. \emph{Individual} fairness seeks to ensure that similar individuals receive comparable outcomes from AI systems~\cite{dwork2012fairness}, while \emph{group} fairness focuses on achieving equitable treatment across demographic groups, such as through metrics like demographic parity~\cite{feldman2015certifying,calders2010three,zafar2017fairnessbeyond}. Further refinements, such as subgroup fairness metrics, aim to address intersections of demographic groups, thereby providing a more nuanced approach \cite{kearns2018preventing}. Meanwhile, causal fairness \cite{kusner2017counterfactual} definitions and techniques like fair representation learning \cite{zemel2013learning} introduce innovative methodologies to address biases within AI systems. Despite these advancements, researchers have highlighted the inherent trade-offs between fairness and accuracy, which pose significant challenges for stakeholders when designing and implementing fair AI systems \cite{kleinberg2016inherent,islam2021free}.

Prior work in algorithmic fairness has increasingly acknowledged that no single fairness metric suffices in real-world contexts, especially when multiple stakeholders are involved \cite{kleinberg2016inherent,mitchell2021algorithmic,narayanan2018translation}. Multi-objective fairness approaches aim to optimize more than one fairness criterion simultaneously, often revealing trade-offs or incompatibilities among them \cite{liu2022accuracy,la2023optimizing}. These efforts have explored Pareto frontiers and weighted optimization schemes to balance competing fairness goals. Meanwhile, a growing body of work has also examined multi-stakeholder fairness, arguing that different societal actors—such as policymakers, service providers, and advocacy groups—bring distinct fairness expectations to AI systems \cite{binns2018fairness}.

A motivating real-world example comes from the Apple Card gender discrimination controversy, where Apple co-founder Steve Wozniak reported receiving a credit limit ten times higher than his wife’s, despite having similar financial profiles.\footnote{www.cnn.com/2019/11/10/business/goldman-sachs-apple-card-discrimination/} This case raised widespread concerns about opaque decision-making and disparate treatment in algorithmic lending. Inspired by this, our framework introduces a fairness metric called \textit{infra-marginal individual fairness}, which compares predicted outcomes between matched individuals who share similar values on a small set of “fair” features (e.g., assets, credit score). This metric captures a merit-based notion of fairness aligned with conservative perspectives, where equitable treatment is expected among similarly qualified individuals, regardless of group membership. Our framework also includes more progressive fairness notions.

A complementary line of work takes a \emph{human-centered} approach to AI fairness, emphasizing participatory processes, stakeholder-facing tools, and documentation to make trade-offs understandable and actionable. Examples include the 2020 workshop on Human-Centered AI Fairness,\footnote{\url{https://fair-ai.academicwebsite.com/}} studies of practitioner needs and pain points \cite{holstein2019improving}, participatory algorithm design with affected communities \cite{lee2019webuildai}, model and data documentation practices \cite{mitchell2019model,gebru2021datasheets}, sociotechnical critiques of abstraction in fairness work \cite{selbst2019fairness}, and organizational checklists for fairness processes \cite{madaio2020co}. These efforts motivate our design choices: ratio-based metrics aligned with legal standards, plain-language explanations, and a multi-stakeholder weighting scheme that lets different actors express and negotiate their priorities.

Although frameworks like that of Berk et al. \cite{berk2017convex} have sought to incorporate \emph{hybrid} approaches combining individual and group fairness, they often do not address the broader societal implications of disparities. These approaches also lack a cohesive, human-centered perspective that systematically integrates diverse fairness considerations and offers stakeholders flexibility in assigning weights to competing objectives. Consequently, there remains a need for comprehensive frameworks that encompass multiple fairness metrics while making trade-offs with predictive performance explicit.

Our framework builds on these human-centered ideas while drawing on two prominent fairness stances—infra-marginality and intersectionality—which offer distinct views on how to treat disparities. We integrate these societal perspectives with AI-specific methods to create a unifying system that (i) supports individual- and group-level assessments, (ii) allows stakeholders to express priorities across outcome vs.\ equality-of-opportunity views, and (iii) makes trade-offs explicit via interpretable weights.

\section{Unifying Fairness Framework}
Ensuring fairness in AI systems requires a holistic approach that incorporates multiple fairness metrics and reflects the diverse priorities of stakeholders~\cite{crawford2019ai}. Our proposed unifying fairness framework is designed to navigate this complexity by enabling stakeholders to assign weights to different fairness metrics based on their priorities. Multiple stakeholders can negotiate fairness weights, helping to counteract developers' ``subjective bias'' \cite{pryzant2020automatically}.

We ground all fairness metrics in the legally inspired disparate impact test, commonly expressed as the 80\% rule~\cite{eeoc1966guidelines} (or more generally, the $p$\%-rule~\cite{zafar2017fairness}). We assume a binary classification problem with features $\mathbf{x}_i \in \mathcal{X}$ and labels $y_i \in \{0,1\}$ for instance $i$. Let $\hat{y}_i \in \{0,1\}$ denote the predicted class label for $i$, and let $\Pr_M(\hat{y}_i{=}1 \mid \mathbf{x}_i)$ denote the model’s predicted probability of the favorable class. For ease of notation, we will sometimes neglect to include the features $\mathbf{x}_i$ in the equations and write $\Pr_M(\hat{y}_i)$ by shorthand when their presence is clear from context. Let $G^+$ and $G^-$ be privileged and unprivileged groups, respectively. A simple ratio-based fairness measure compares group averages:
\begin{equation}
\mathcal{R} \;=\; \frac{\displaystyle \text{avg}_{i \in G^-}\, q(i)}{\displaystyle \text{avg}_{j \in G^+}\, q(j)} ,
\label{eq:base_ratio_simple}
\end{equation}
where $i$ indexes individuals in the unprivileged group $G^-$ and $j$ indexes individuals in the privileged group $G^+$. Here $q(\cdot)$ is the quantity being compared (e.g., $q(i)=\Pr_M(\hat{y}_i)$, or $q(i)=\Pr_M(\hat{y}_i \mid y_i=1)$ for EOO, with $y{=}1$ indicating the positive ground-truth class). Values of $\mathcal{R}$ closer to 1 indicate greater parity; a threshold of $0.8$ instantiates the 80\% rule~\cite{eeoc1966guidelines}.

To formalize fairness, our framework supports eight distinct fairness metrics that span multiple normative perspectives. These metrics arise from combining four core fairness notions—\emph{individual vs.\ group} and \emph{infra-marginal vs.\ intersectional}—with two evaluation regimes: \emph{outcome-based} and \emph{equality-of-opportunity (EOO)-based}. Rather than privileging one definition, our unified formulation allows flexible trade-offs, enabling stakeholders to assign and negotiate weights based on their specific fairness goals. All eight metrics use a consistent and straightforward ratio-based formulation to make the metrics and their relationships easy to understand for non-expert stakeholders.

\subsection{Fairness Metrics}
We instantiate the simple ratio in Eq.~\eqref{eq:base_ratio_simple} along three dimensions:

\begin{enumerate}
\item \textbf{Individual vs.\ Group.}
Individual fairness compares people one-to-one; group fairness compares averages across groups. Starting from Eq.~\eqref{eq:base_ratio_simple}, the \emph{group} version uses group averages as written. To make it \emph{individual}, we compare cross-group \emph{pairs} of people who are similarly qualified (matched on $r^{(f)}$), and then average those pairwise ratios. Pairwise comparisons ensure a comparison between similarly situated individuals, so differences are not due to qualification gaps (e.g., two applicants with similar creditworthiness should not get systematically different predictions).

\item \textbf{Infra-marginal vs.\ Intersectional.}
Infra-marginality assumes a level-playing-field baseline and compares only people with similar estimated merit; intersectional emphasizes systemic inequities across demographics and their intersections. For \emph{infra-marginal}, we require similarity on the fair-risk score $r^{(f)}$ (see Eq.~\eqref{eq:risk_score}) before comparing. For \emph{intersectional}, we \emph{do not} require matching at the group level; for the individual case, we first align baselines by a simple constant shift for the unprivileged group and then form pairs in that adjusted space. Matching (akin to propensity score matching \cite{rosenbaum1985constructing}) isolates model behavior at a fixed ``merit'' level; the parity shift in the individual–intersectional case explicitly counters baseline disadvantage before making a comparison at fixed merit. Our matching approach is inspired by the Apple Card controversy, in which unfairness was alleged when similarly qualified individuals of differing gender (spouses with identical assets and similar credit scores) were assigned vastly different outcomes by an AI system.

\item \textbf{Outcome vs.\ Equality of Opportunity (EOO).}
\emph{Outcome parity} aims for parity between sensitive demographics across all predictions~\cite{dwork2012fairness}; \emph{equality of opportunity (EOO)} fairness focuses on those who truly qualify~\cite{hardt2016equality}. We keep the same ratio structure and simply change what we average: $q(i)=\Pr_M(\hat{y}_i)$ for \emph{Outcome}, vs.\ $q(i)=\Pr_M(\hat{y}_i\mid y_i=1)$ for \emph{EOO}~\cite{hardt2016equality}. Outcome parity asks whether predictions are balanced overall; EOO asks whether qualified individuals are treated comparably.
\end{enumerate}

\noindent We now give the concise equations that implement these switches.

\textbf{Fair-risk scores for matching.} When matching is required (i.e., in the infra-marginal variants), we use a simple fair-risk score,
\begin{equation}
    r^{(f)} \;\triangleq\; \Pr(\hat{y}=1 \mid \mathbf{x}^{(f)}).
    \label{eq:risk_score}
\end{equation}
estimated with a basic model (e.g., logistic regression) on a small set of relatively ``fair'' features $\mathbf{x}^{(f)}$ (we use one per dataset in our experiments; see Section~\ref{sec:exp_setup}). For instance, in the Apple Card scenario, the ``fair features'' would be assets and credit scores. We match people across groups when $r^{(f)}$ is similar (or after the constant shift above for the individual–intersectional variant), so comparisons reflect model differences rather than underlying qualification gaps.

\textbf{(A) Granularity — Individual vs.\ Group.}
Eq.~\eqref{eq:base_ratio_simple} is a \emph{group} ratio. To make it \emph{individual}, we replace group averages by an average of \emph{pairwise ratios} across matched cross-group pairs $(i,j)$ with similar $r^{(f)}$:
\[
\text{Outcome:}\quad 
\mathcal{R}_{\text{ind}} \;=\; \text{avg}_{(i,j)\,\text{matched on } r^{(f)}} \!\Bigg[\frac{\Pr_M(\hat{y}_i)}{\Pr_M(\hat{y}_j)}\Bigg],
\qquad
\text{EOO:}\quad 
\mathcal{R}_{\text{ind}}^{\text{EOO}} \;=\; \text{avg}_{(i,j)} \!\Bigg[\frac{\Pr_M(\hat{y}_i\mid y_i=1)}{\Pr_M(\hat{y}_j\mid y_j=1)}\Bigg].
\]
This enforces a comparison at fixed merit $r^{(f)}$, so differences reflect model behavior rather than qualification gaps. We construct cross-group matches using nearest-neighbor matching on the scalar fair-risk score $r^{(f)}$ estimated from the fair features. Concretely, we designate the smaller of the two groups (by sample size) as the query set and the larger group as the reference set. For each query instance $i$, we select its single nearest neighbor $j$ in the reference group by minimizing $|r^{(f)}_i - r^{(f)}_j|$. This procedure may map multiple query instances to the same reference instance (i.e., many-to-one). This simple 1-NN propensity-score-style matching reduces imbalance on the matching dimension \cite{rosenbaum1983central,rosenbaum1985constructing}.

\textbf{(B) Societal stance — Infra-marginal vs.\ Intersectional.}
For \emph{infra-marginal}, we keep the “similar on $r^{(f)}$” requirement (see Eq.~\eqref{eq:risk_score}) for individuals (pairs) and, at group level, we focus on people with similar $r^{(f)}$ levels across groups.
For \emph{intersectional}, we drop matching entirely for groups; for individuals, we first apply a constant shift so the unprivileged average equals the privileged average, then match in that adjusted space:
\[
r^{(f)*}_j \;=\; r^{(f)}_j + c,
\qquad
c \;=\; \text{avg}_{i\in G^-}\!\big[r^{(f)}_i\big] - \text{avg}_{j\in G^+}\!\big[r^{(f)}_j\big],
\]
and use the same pairwise ratios as above but with $r^{(f)*}$ used for matching. Infra-marginal isolates model effects at fixed merit; the individual–intersectional shift levels baselines before enforcing a merit-controlled comparison, thus implementing a form of ``affirmative action.''

\textbf{(C) Evaluation regime — Outcome vs.\ EOO.}
We set $q(i)=\Pr_M(\hat{y}_i)$ for \emph{Outcome} or $q(i)=\Pr_M(\hat{y}_i\mid y_i=1)$ for \emph{EOO} in the same formulas, without any further structural change. Outcome parity measures overall balance; EOO focuses on qualified cases (e.g., equalizing true positives).

\vspace{0.5em}
Putting (A)–(C) together yields the eight metrics used in our framework. Table~\ref{tab:unifyingFramework} summarizes the eight metrics from the three toggles above, using the same ratio idea with the appropriate scope (pairs vs.\ groups) and conditioning (Outcome vs.\ EOO). Rows set the granularity (individual vs.\ group) and columns set the stance (infra-marginal uses matching on $r^{(f)}$; intersectional does not).

\begin{table}[t]
\footnotesize
\setlength{\tabcolsep}{4pt}
\renewcommand{\arraystretch}{1.25}
\centering
\begin{tabular}{@{} c c c c @{}}
\toprule
& \multicolumn{2}{c}{\textbf{Infra-marginal}} & \textbf{Intersectional} \\
\cmidrule(lr){2-3}\cmidrule(lr){4-4}

\multirow{3}{*}{\textbf{Individual}}
  & \multicolumn{2}{c}{\colorbox{gray!15}{\strut Matched pairs: $r_i^{(f)} \approx r_j^{(f)}$}}
  & \colorbox{gray!15}{\strut Matched pairs (parity-adjusted $r^{(f)}$)} \\
\cmidrule(lr){2-3}\cmidrule(lr){4-4}
  & Outcome
  & $avg(\Pr_M(\hat{y}_i) / \Pr_M(\hat{y}_j))$
  & $avg(\Pr_M(\hat{y}_i) / \Pr_M(\hat{y}_j))$ \\
\cmidrule(lr){2-3}\cmidrule(lr){4-4}
  & Equality of Opportunity
  & $avg(\Pr_M(\hat{y}_i \mid y_i=1) / \Pr_M(\hat{y}_j \mid y_j=1))$
  & $avg(\Pr_M(\hat{y}_i \mid y_i=1) / \Pr_M(\hat{y}_j \mid y_j=1))$ \\
\midrule

\multirow{3}{*}{\textbf{Group}}
  & \multicolumn{2}{c}{\colorbox{gray!15}{\strut Similar $r^{(f)}$ levels across groups}}
  & \colorbox{gray!15}{\strut Unmatched intersectional subgroups} \\
\cmidrule(lr){2-3}\cmidrule(lr){4-4}
  & Outcome
  & $avg(\Pr_M(\hat{y}_i)) / avg(\Pr_M(\hat{y}_j))$
  & $avg(\Pr_M(\hat{y}_i)) / avg(\Pr_M(\hat{y}_j))$ \\
\cmidrule(lr){2-3}\cmidrule(lr){4-4}
  & Equality of Opportunity
  & $avg(\Pr_M(\hat{y}_i \mid y_i=1)) / avg(\Pr_M(\hat{y}_j \mid y_j=1))$
  & $avg(\Pr_M(\hat{y}_i \mid y_i=1)) / avg(\Pr_M(\hat{y}_j \mid y_j=1))$ \\
\bottomrule
\end{tabular}
\caption{\footnotesize The proposed framework systematically encodes fairness definitions across levels of granularity and value systems. Here, $i$ and $j$ refer to instances from different demographic groups (e.g., male, female), and $avg(\cdot)$ computes the mean. Two perspectives are considered: \emph{Outcome}, which measures fairness based on model predictions alone, and \emph{Equality of Opportunity}, which evaluates predictions conditioned on true positives ($y = 1$), aligning with \cite{hardt2016equality}. \textbf{Matched pairs means we only compare two people from different groups if they have similar fair-risk scores $r^{(f)}$.} Ratios of predicted probabilities are analogous to the 80\% rule \cite{eeoc1966guidelines}.}
\label{tab:unifyingFramework}
\end{table}

\subsection{Multi-Stakeholder, Multi-Objective Optimization} \label{sec:methodology}
This section focuses on the multi-objective, multi-stakeholder part of our method: we treat accuracy and the eight ratio-based fairness metrics as competing goals and let stakeholders combine them with interpretable weights. 

Let $R_1,\ldots,R_8$ be the eight fairness metrics from Table~\ref{tab:unifyingFramework}. We assign nonnegative weights $w_m$ (optionally summing to 1) to form the overall fairness term:
\begin{equation}
  R(\mathbf{X}; \theta) \;=\; \sum_{m=1}^{8} w_m \, R_m(\mathbf{X}; \theta).
  \label{eqn:weight_matrix}
\end{equation}
We then trade off accuracy and fairness via:
\begin{equation}
    \min_{\theta}\; f(\mathbf{X}; \theta) 
    \;\triangleq\; \frac{1}{N}\sum_{n=1}^{N} L(\mathbf{x}_n; \theta) \;-\; \lambda \, R(\mathbf{X}; \theta),
    \label{eqn:objectivefunction}
\end{equation}
where $L$ is the primary prediction loss (e.g., binary cross-entropy) and $\lambda \ge 0$ controls the overall fairness–performance balance.

To visualize stakeholder negotiations, in our experiments we often activate only two metrics $R_a$ and $R_b$:
\begin{equation}
R(\mathbf{X}; \theta) \;=\; \alpha \, R_a(\mathbf{X}; \theta) + (1-\alpha)\, R_b(\mathbf{X}; \theta),
\qquad \alpha \in [0,1],
\label{eq:pairwise_fairness}
\end{equation}
with the remaining weights set to zero. Sweeping $\alpha$ (for a fixed $\lambda$) traces fairness–fairness curves, showing how improving one notion typically requires sacrificing the other; jointly varying $\alpha$ and $\lambda$ maps out fairness–accuracy–fairness frontiers \cite{liu2022accuracy,la2023optimizing}. This helps select settings that reflect stakeholder values.

We optimize the overall objective in Eq.~\eqref{eqn:objectivefunction} with Adam (learning rate $1{\times}10^{-4}$) in PyTorch. Each epoch computes the \emph{full-batch} objective on the entire training set (no mini-batches). The primary loss $L$ is binary cross-entropy applied to the model’s sigmoid probabilities, and the overall objective includes the weighted fairness term from Eq.~\eqref{eqn:weight_matrix}. We train for 2000 epochs and evaluate on a held-out development set after every epoch. Full-batch updates ensure matching-based fairness metrics are computed consistently on the whole dataset at each optimization step.

\section{Experimental Setup}
\label{sec:exp_setup}

\begin{figure}[tb]
    \centering
    \begin{subfigure}[b]{0.45\textwidth}
        \centering
        \includegraphics[width=\textwidth]{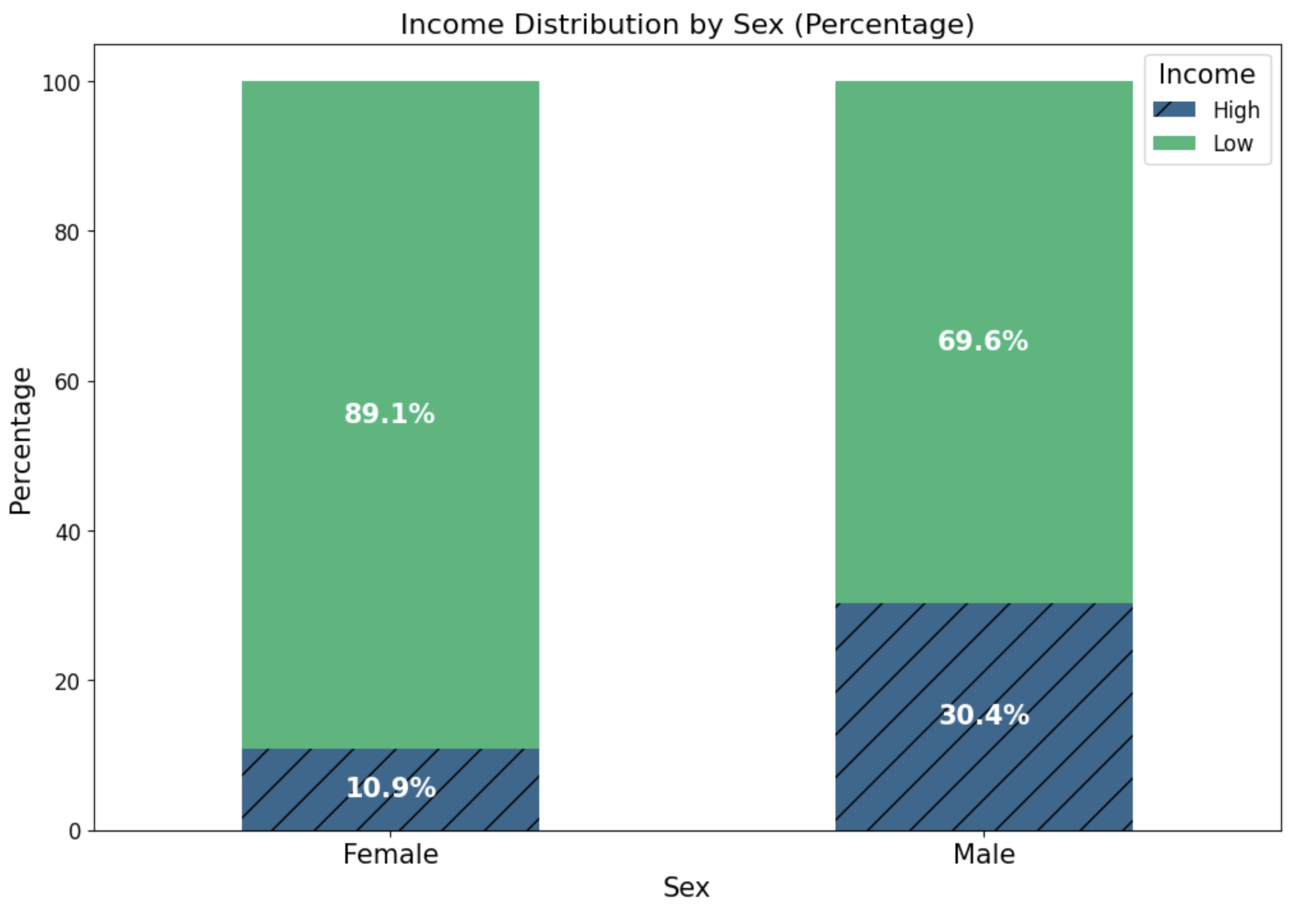}
        \caption{UCI Adult Dataset}
        \label{fig:adult_data}
    \end{subfigure}
    \hspace{0.02\textwidth} 
    \begin{subfigure}[b]{0.45\textwidth}
        \centering
        \includegraphics[width=\textwidth]{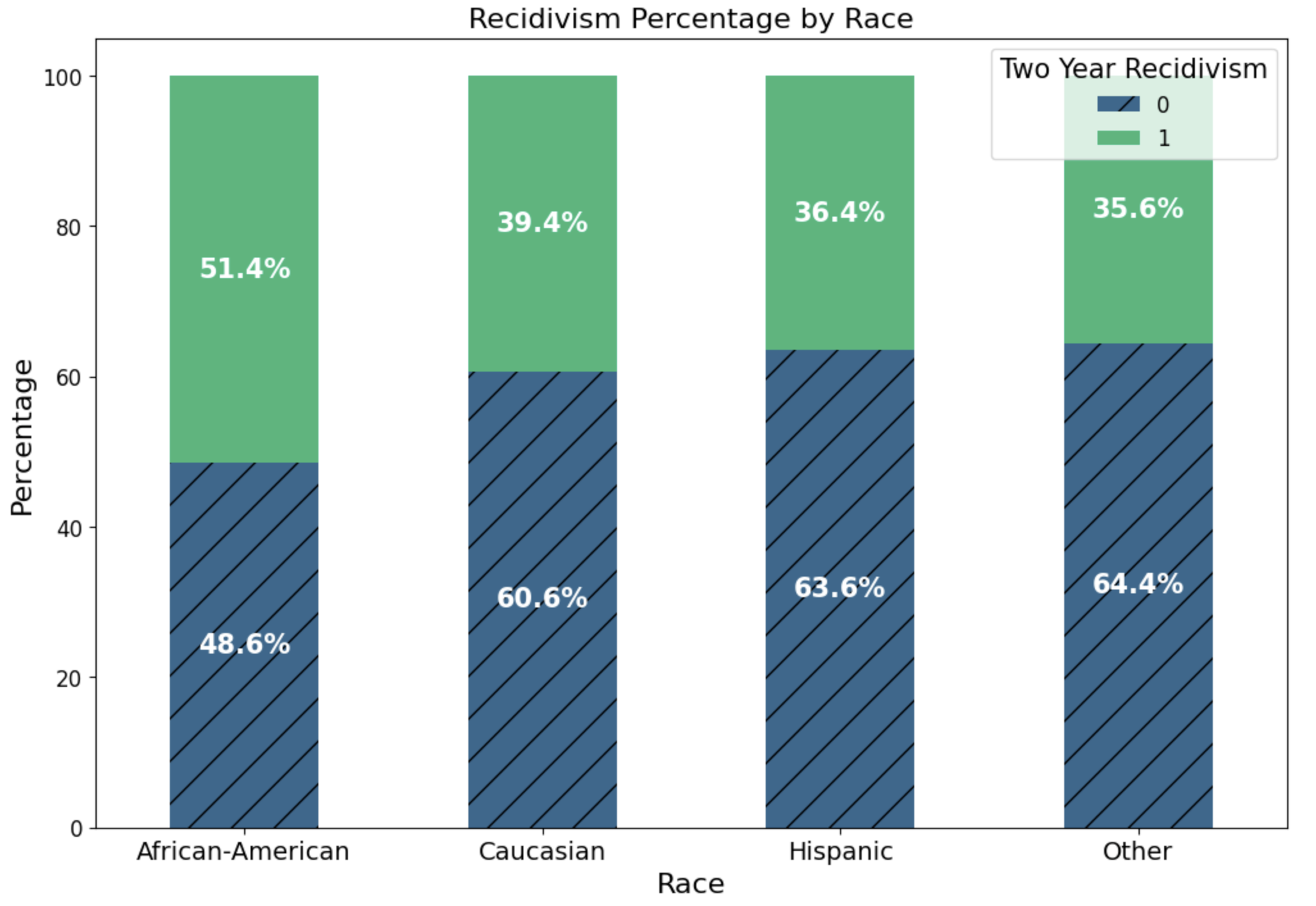}
        \caption{ProPublica COMPAS Dataset}
        \label{fig:compas_data}
    \end{subfigure}
    
    \vspace{0.02\textwidth} 

    \begin{subfigure}[b]{0.45\textwidth}
        \centering
        \includegraphics[width=\textwidth]{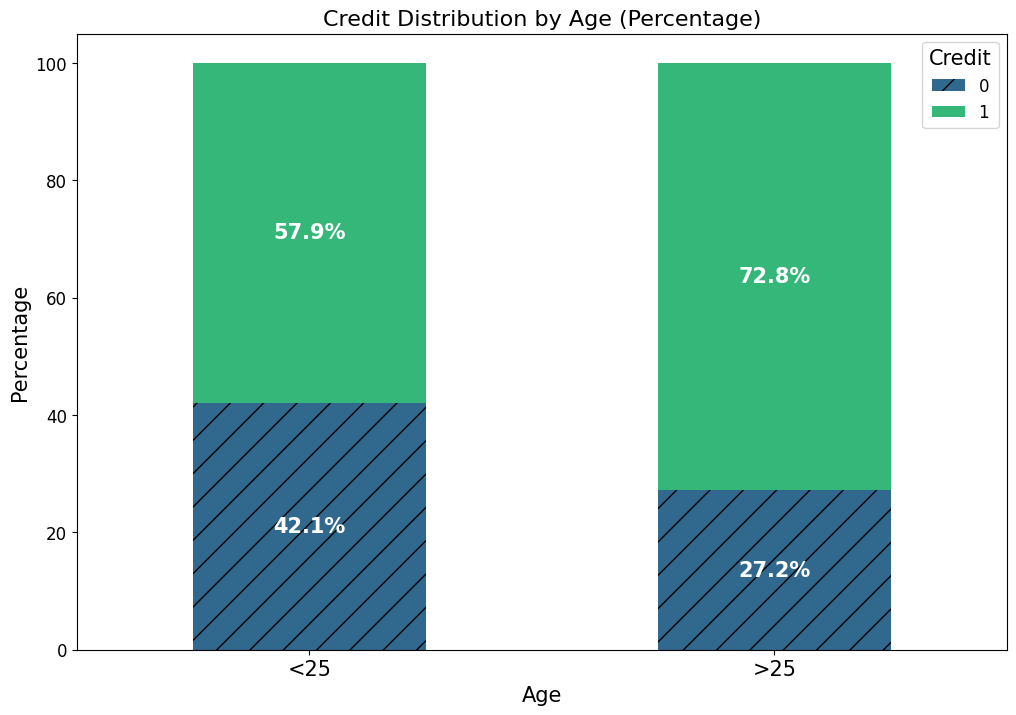}
        \caption{German Credit Dataset}
        \label{fig:german_data}
    \end{subfigure}
    \hspace{0.02\textwidth} 
    \begin{subfigure}[b]{0.45\textwidth}
        \centering
        \includegraphics[width=\textwidth]{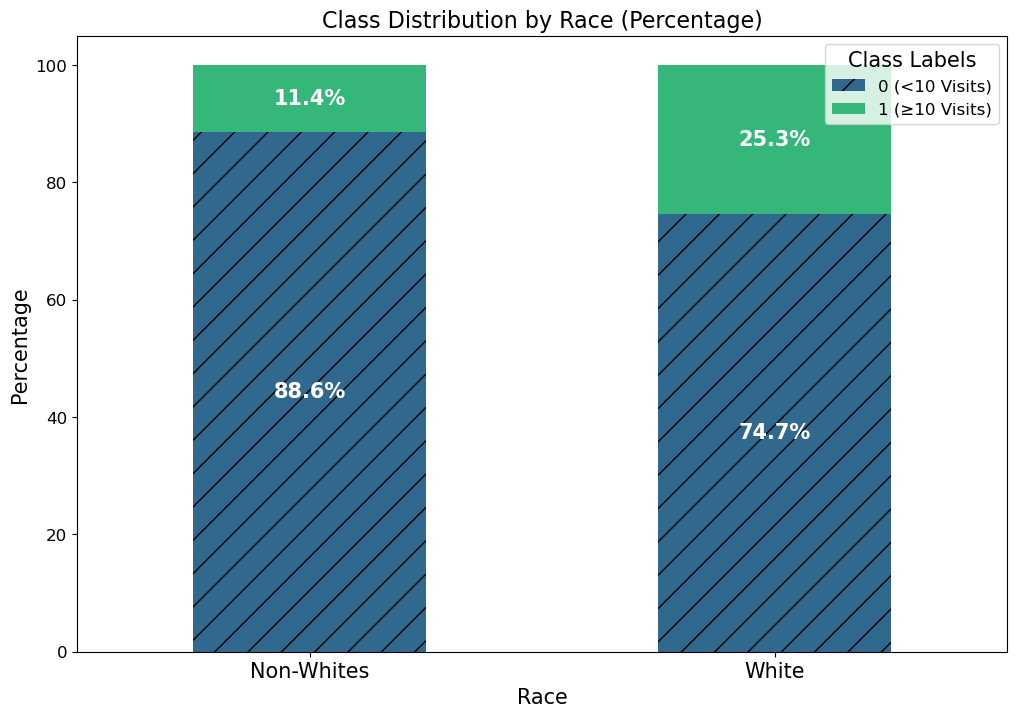}
        \caption{MEPS Dataset}
        \label{fig:meps_data}
    \end{subfigure}
    \caption{Class distribution across protected groups in the Adult, COMPAS, MEPS, and German Credit datasets. For UCI Adult, income $\leq\$50K$ is labeled as ``Low'' and $>\$50K$ as ``High.'' For COMPAS, ``1'' is the positive label, indicating a prediction of recidivism. For MEPS, the positive class represents high medical utilization (10 or more visits per year). In German Credit, ``1'' denotes creditworthiness and ``0'' indicates credit denial, with noticeably higher denial rates among younger applicants. These disparities highlight the presence of label imbalance and potential bias across demographic subgroups, underscoring the need for fairness-aware modeling.}
    \label{fig:class_distr_data_sets}
\end{figure}

We evaluated our framework using four benchmark datasets commonly employed in fairness research: the UCI Adult Census dataset~\cite{misc_adult_2,kelly2023uci}, the COMPAS recidivism dataset~\cite{angwin2016machine}, the Medical Expenditure Panel Survey (MEPS\footnote{https://meps.ahrq.gov/mepsweb/}) dataset \cite{bellamy2018ai}, and the German Credit dataset \cite{statlog_german_credit_data_144}. Figure \ref{fig:class_distr_data_sets} shows the class distribution across different protected attributes. These datasets are widely recognized for assessing fairness in machine learning tasks across domains such as income prediction, criminal justice, utilization for medical care, and credit risk assessment.

All experiments use a feed-forward neural network with a single hidden layer of 64 units: Linear $\rightarrow$ ReLU $\rightarrow$ Linear $\rightarrow$ Sigmoid (1-D output). The same architecture is used for all datasets for comparability. Training details (objective, loss, and optimization) are given in Sec.~\ref{sec:methodology}.

To compute the “fair risk score” for each individual, we selected baseline features considered relatively fair within each dataset. Specifically, we used education level for the UCI Adult dataset, prior offenses count for COMPAS, the Physical Component Summary (PCS) score for MEPS, and credit history for the German Credit dataset. We then trained logistic regression models on these baseline features to estimate fair risk scores.\\

\noindent\textbf{Dataset Descriptions:}
The \textbf{UCI Adult} dataset contains 48,842 instances, each described by 15 attributes (6 numerical, 7 categorical, and 2 binary). The task is to predict whether an individual earns more than \$50,000 per year. We treat ``Sex'' as the protected attribute. The \textbf{COMPAS} dataset, compiled by ProPublica \cite{angwin2016machine}, includes information on 7,214 defendants and is used to predict the likelihood of criminal recidivism. The binary target indicates whether a defendant will re-offend within two years. We use ``Race'' as the protected attribute, binarizing it to compare ``Caucasian'' vs. ``African-American'' only. Original race labels such as ``Asian,'' ``Native American,'' and ``Other'' were merged due to small sample sizes. The COMPAS system has been criticized for amplifying racial bias in risk assessments. The \textbf{MEPS} (Medical Expenditure Panel Survey) dataset is a nationally representative survey of healthcare utilization and expenditures in the U.S. Using the AIF360 implementation~\cite{bellamy2018ai}, we extract demographic and health-related features such as age, gender, race, education, insurance status, and number of medical visits. The prediction task is to identify whether an individual has high medical utilization (10 or more visits in a year). We define ``Race'' (White vs. Non-White) as the protected attribute. The \textbf{German Credit} dataset contains 1,000 instances with 20 attributes describing financial status, credit history, and demographic factors. The task is to predict whether an applicant is creditworthy. Following prior work, we binarize the protected attribute ``Age'' into ``Young'' ($\leq 25$) and ``Not Young'' ($> 25$)~\cite{bellamy2018ai,friedler2019comparative}. This dataset is commonly used in fairness research to study disparities in credit approval outcomes.

All four datasets exhibit disparities in positive outcome rates across demographic groups (see Figure~\ref{fig:class_distr_data_sets}), motivating fairness-aware modeling. 

\section{Results}
In this section, we evaluate how our unified fairness framework performs across multiple datasets and fairness configurations. We analyze both fairness-accuracy trade-offs and fairness-fairness tensions by varying the regularization strength (\( \lambda \)) and the relative weights assigned to different fairness metrics. The results offer insight into the flexibility and effectiveness of our approach in modeling stakeholder-driven fairness objectives across diverse application domains.

\begin{figure}[p]
    \centering

    \scalebox{0.82}{ 
    \begin{subfigure}[b]{\textwidth}
        \centering
        \begin{subfigure}[b]{0.45\textwidth}
            \centering
            \includegraphics[width=\textwidth]{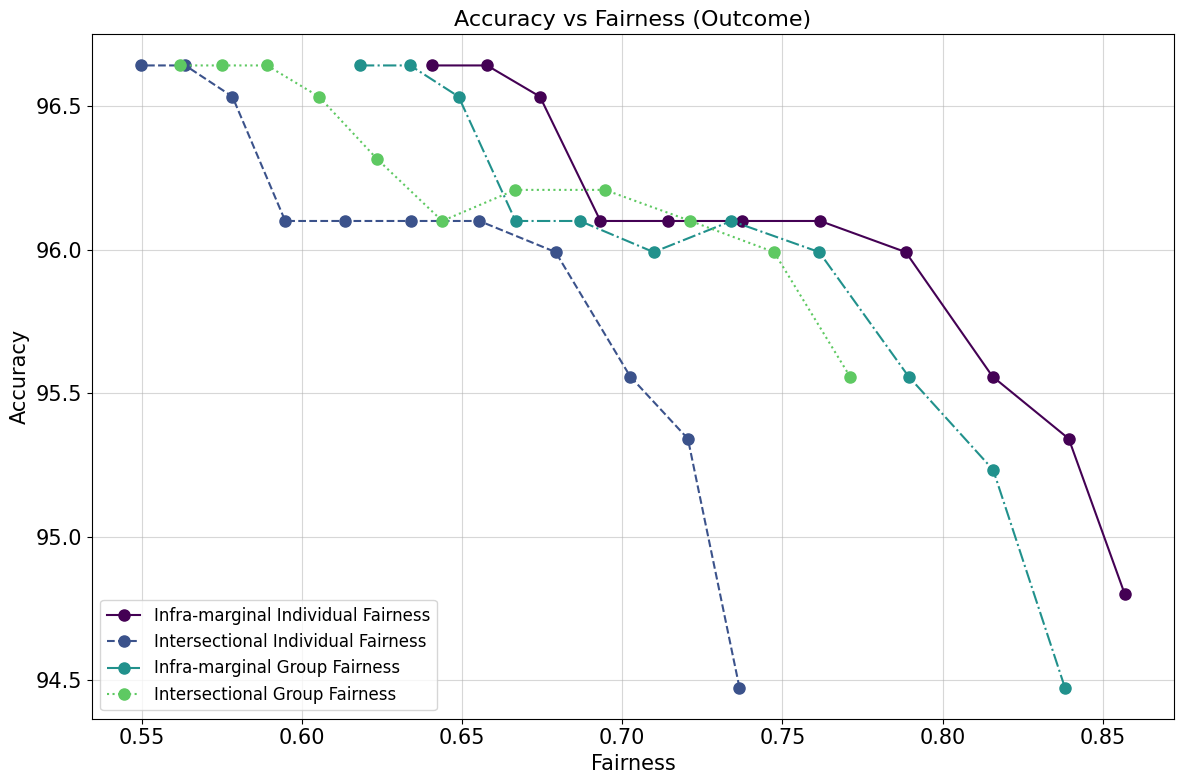}
        \end{subfigure}
        \hfill
        \begin{subfigure}[b]{0.45\textwidth}
            \centering
            \includegraphics[width=\textwidth]{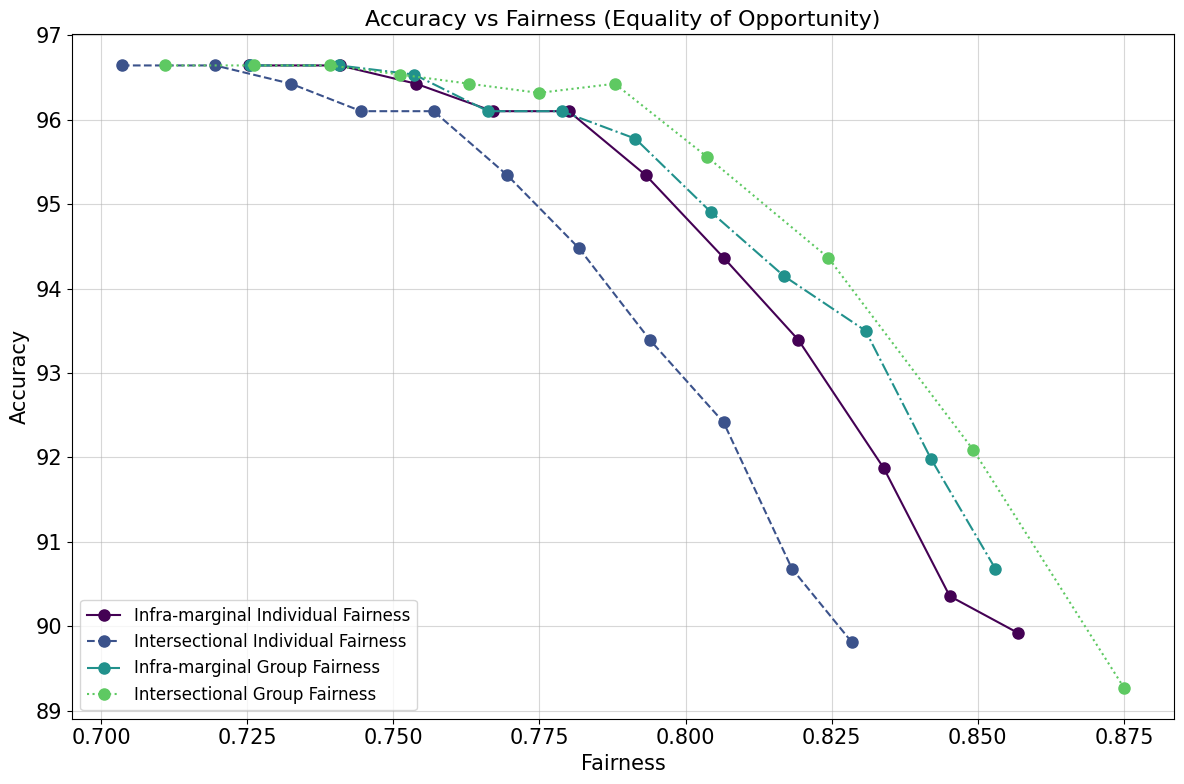}
        \end{subfigure}
        \caption{ProPublica COMPAS Dataset}
        \label{fig:accuracy_fairness_compas}
    \end{subfigure}
    }

    \vspace{1em}

    \scalebox{0.82}{ 
    \begin{subfigure}[b]{\textwidth}
        \centering
        \begin{subfigure}[b]{0.45\textwidth}
            \centering
            \includegraphics[width=\textwidth]{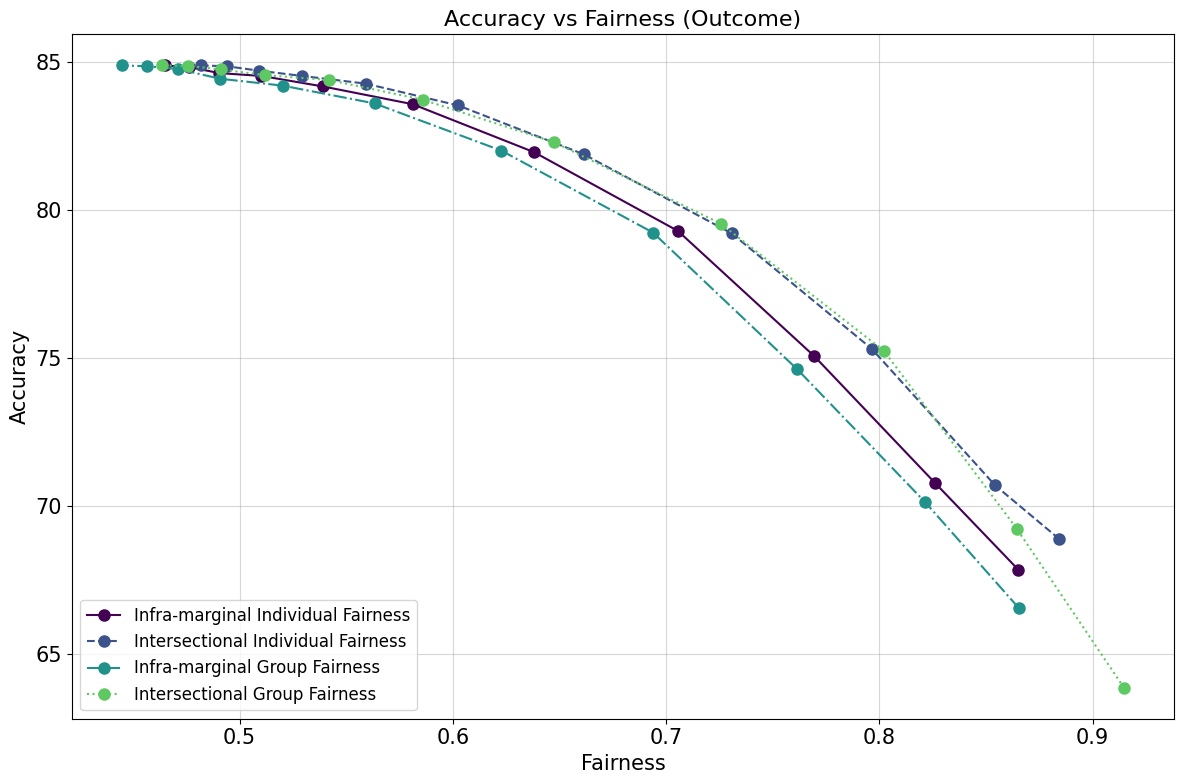}
        \end{subfigure}
        \hfill
        \begin{subfigure}[b]{0.45\textwidth}
            \centering
            \includegraphics[width=\textwidth]{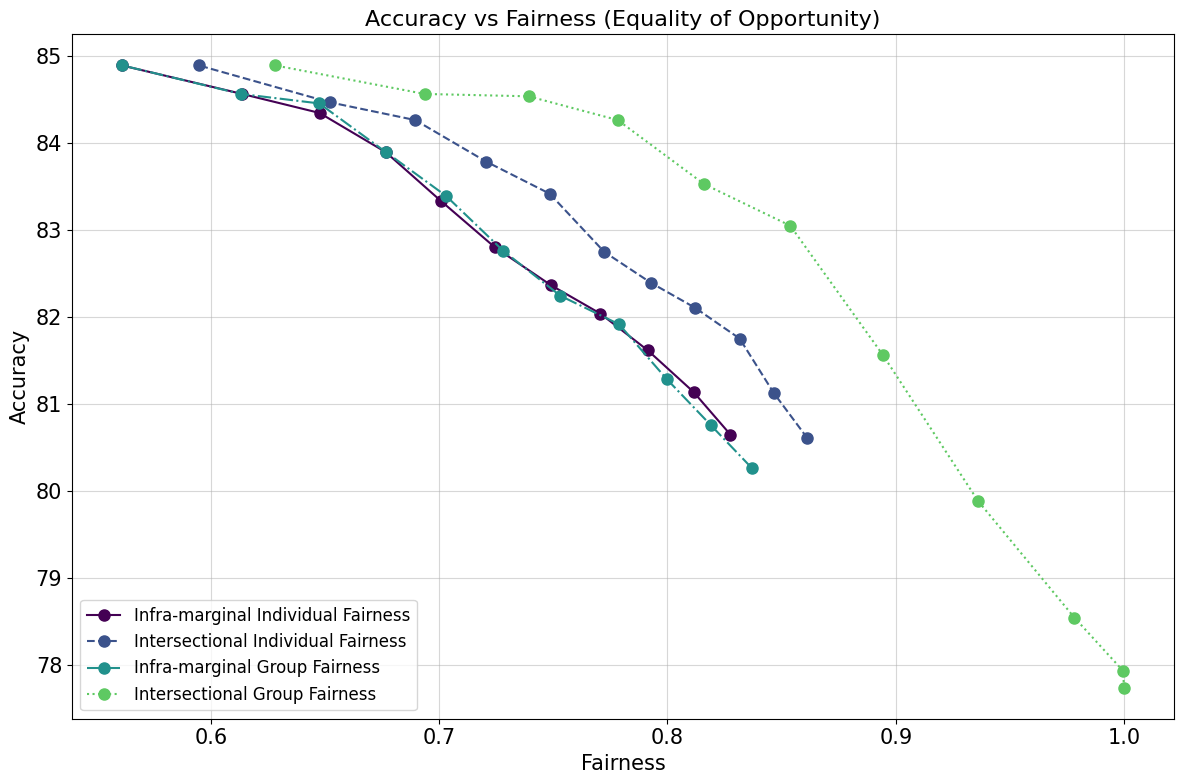}
        \end{subfigure}
        \caption{UCI Adult Dataset}
        \label{fig:accuracy_fairness_adult}
    \end{subfigure}
    }

    \vspace{1em}

    \scalebox{0.82}{ 
    \begin{subfigure}[b]{\textwidth}
        \centering
        \begin{subfigure}[b]{0.45\textwidth}
            \centering
            \includegraphics[width=\textwidth]{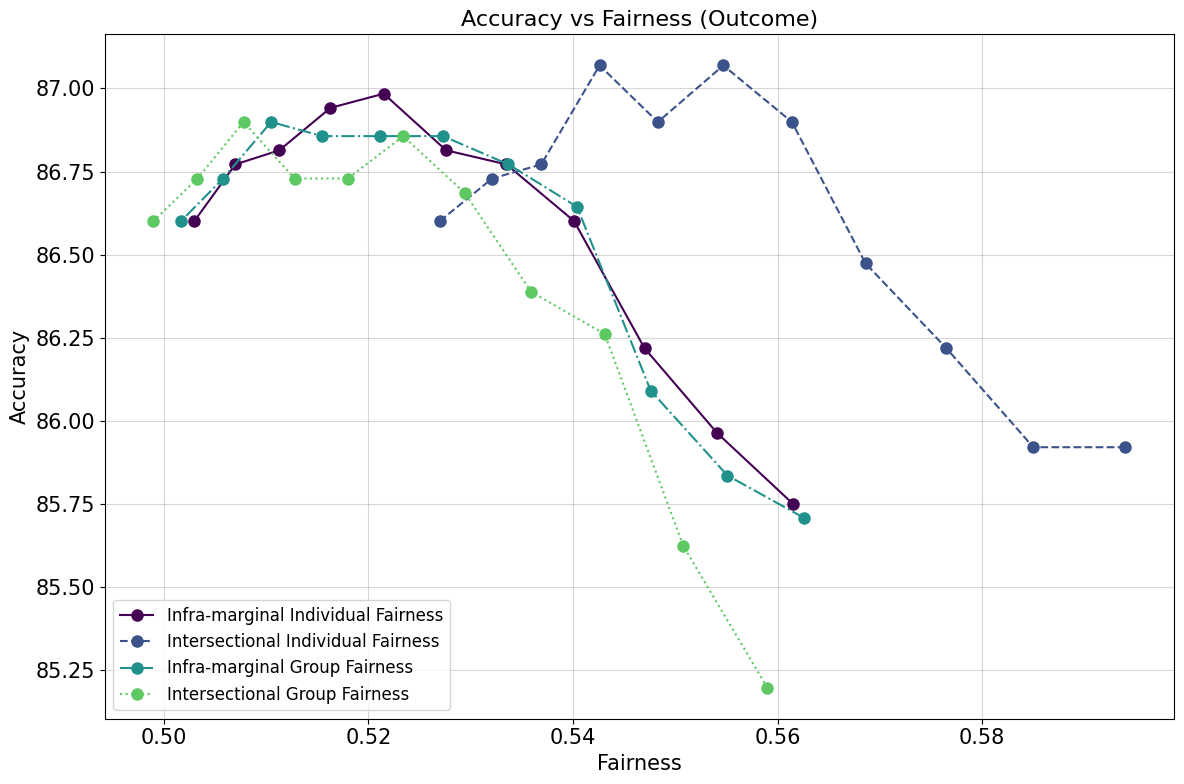}
        \end{subfigure}
        \hfill
        \begin{subfigure}[b]{0.45\textwidth}
            \centering
            \includegraphics[width=\textwidth]{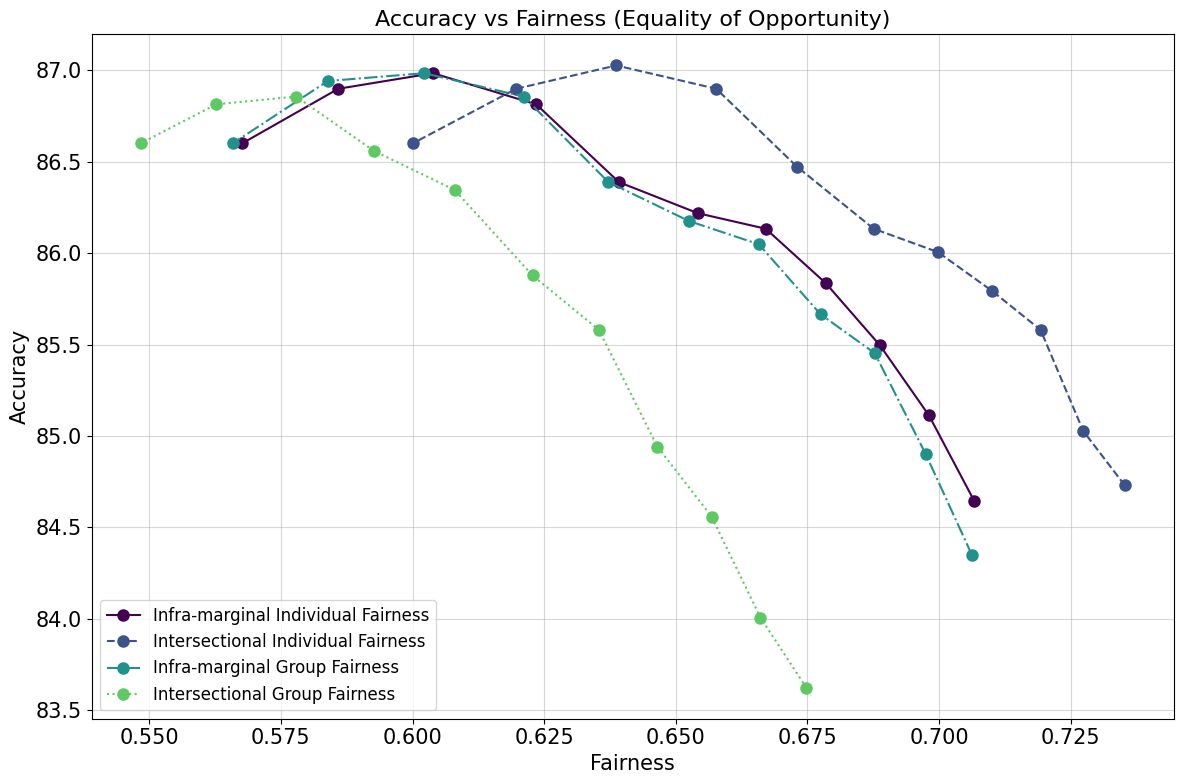}
        \end{subfigure}
        \caption{MEPS Dataset}
        \label{fig:accuracy_fairness_meps}
    \end{subfigure}
    }

    \scalebox{0.82}{ 
    \begin{subfigure}[b]{\textwidth}
        \centering
        \begin{subfigure}[b]{0.45\textwidth}
            \centering
            \includegraphics[width=\textwidth]{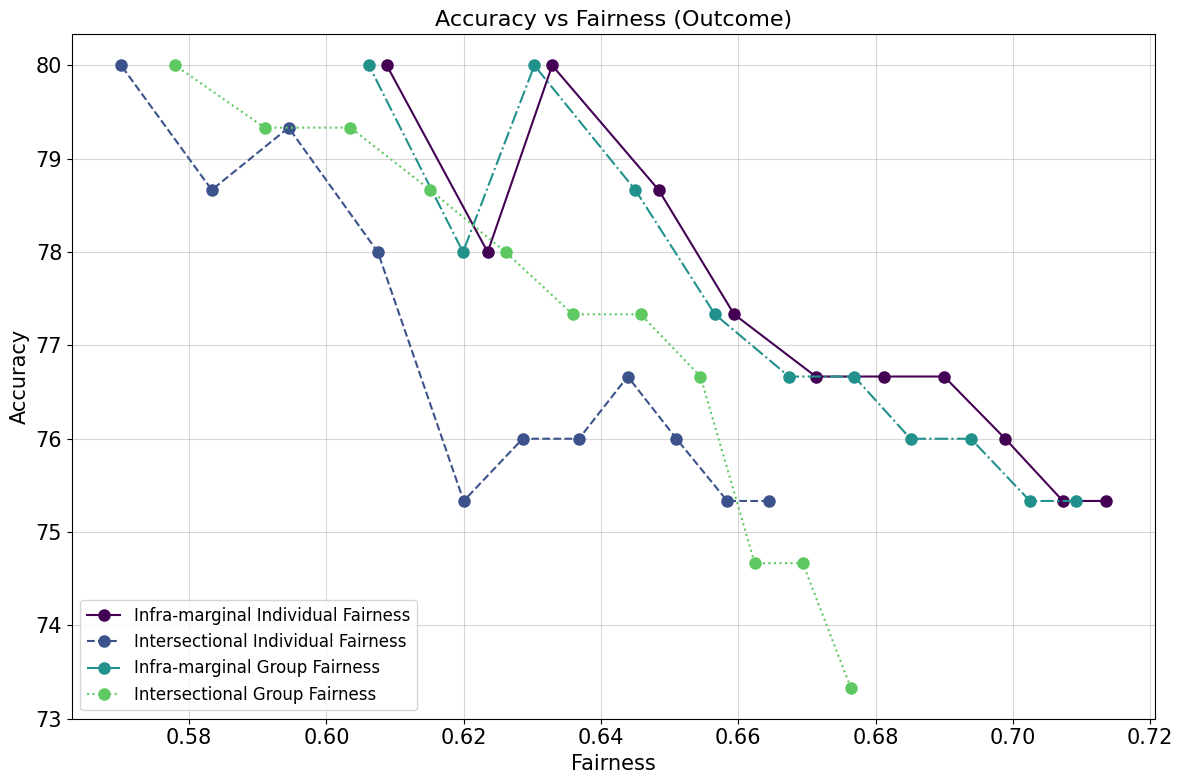}
        \end{subfigure}
        \hfill
        \begin{subfigure}[b]{0.45\textwidth}
            \centering
            \includegraphics[width=\textwidth]{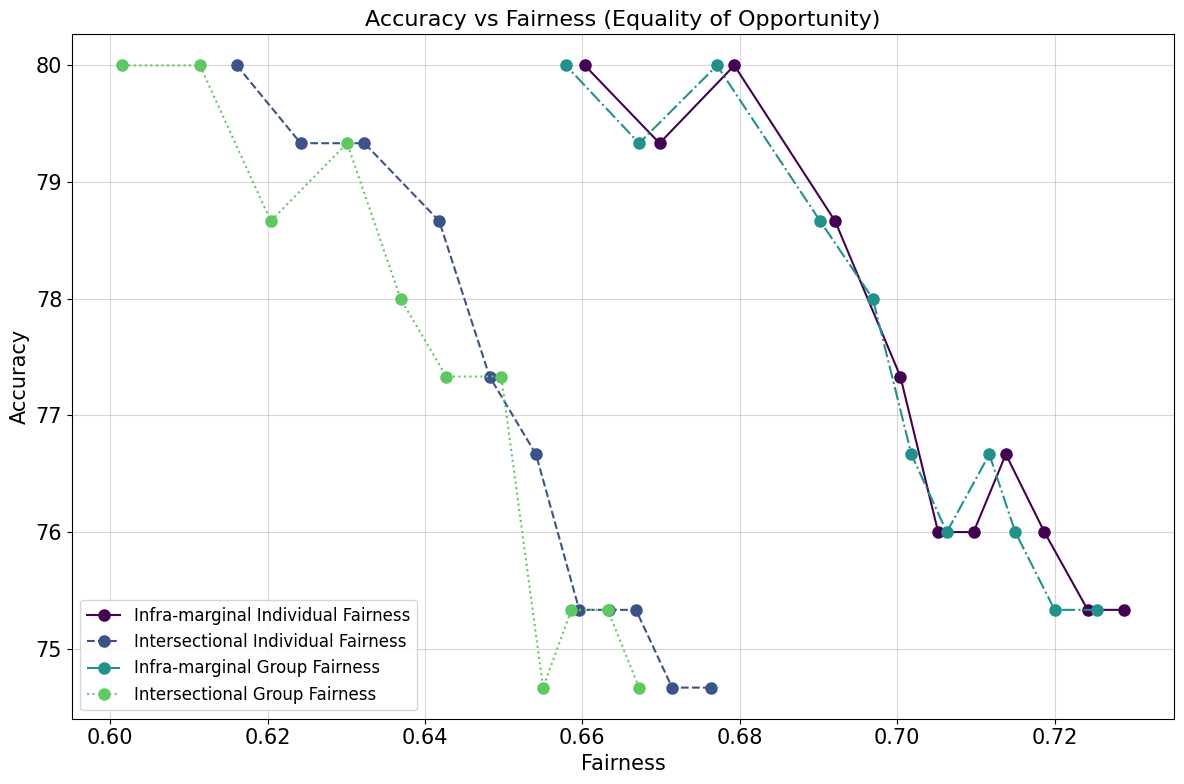}
        \end{subfigure}
        \caption{German Credit Dataset}
        \label{fig:accuracy_fairness_german}
    \end{subfigure}
    }

    \caption{Trade-off between accuracy and fairness across datasets and fairness definitions. Each data-point represents the trade-off for different $\lambda$.}
    \label{fig:accuracy_fairness_tradeoff}
\end{figure}

\subsection{Accuracy-Fairness Trade-Off}
In this section, we examine the relationship between predictive performance and fairness by varying the regularization parameter \( \lambda \), which controls the strength of fairness enforcement. Figure~\ref{fig:accuracy_fairness_tradeoff} presents accuracy–fairness curves for all four datasets: UCI Adult, COMPAS, MEPS, and German Credit. Each point corresponds to a specific \( \lambda \in [0, 0.1, 0.2, \ldots, 0.9, 1] \), where higher \( \lambda \) values place greater emphasis on fairness. The case \( \lambda = 0 \) corresponds to the \emph{typical model}, which is trained without explicit fairness constraints and serves as the baseline for accuracy and fairness comparisons.

Across datasets, we observe the expected trend: increasing fairness generally comes at the expense of predictive accuracy. The magnitude of this trade-off depends on how far a metric’s target fairness is from the typical model’s baseline. Metrics that begin closer to their target require smaller adjustments and therefore tend to preserve accuracy better, while those starting further away demand larger corrective shifts that result in greater accuracy losses.

\begin{itemize}
    \item \textit{COMPAS Dataset:}  
    In both outcome-based (left) and EOO-based (right) settings, the curves begin at \( \lambda = 0 \) with the typical model’s baseline accuracy (\(\approx 96.64\%\)) and fairness for each metric. Baseline fairness varies notably across metrics—ranging from about 0.55 to 0.64 in the outcome-based case, and from about 0.70 to 0.73 in the EOO-based case. 
    Although the shapes of the curves differ slightly between outcome-based and EOO-based fairness, both show the same general pattern: improvements in fairness come with accuracy costs, and the magnitude of those costs depends on the metric’s starting point in the typical model.\\

    \item \textit{Adult Dataset:}  
    For the Adult dataset, baseline accuracy is \(\approx 84.89\%\), with baseline fairness ranging from 0.46–0.49 (outcome-based) and 0.56–0.63 (EOO-based). Accuracy remains relatively stable at lower \( \lambda \) values, with more noticeable declines as fairness is emphasized further. While the shapes of the curves differ slightly between outcome-based and EOO-based fairness, both follow the same general trend: increasing fairness is accompanied by accuracy costs, and the magnitude of these costs depends on how far the baseline fairness of a metric is from higher fairness levels.\\

    \item \textit{MEPS Dataset:}  
    The MEPS dataset shows relatively balanced starting points, with baseline fairness between 0.50–0.56 (outcome-based) and 0.55–0.71 (EOO-based), alongside a baseline accuracy of \(\approx 86.6\%\). Although accuracy remains relatively stable for small to moderate \( \lambda \), larger fairness gains at higher \( \lambda \) values are accompanied by visible drops in accuracy. Both fairness definitions show the same general trend: moving further from the typical model’s baseline fairness tends to come at an increasing cost to predictive performance. A slight initial rise in accuracy at low \( \lambda \) is also visible in both cases, consistent with the ``fairness for free'' effect~\cite{islam2021free}, though the magnitude here is modest.\\

    \item \textit{German Credit Dataset:}  
    The German Credit dataset, with only 1,000 samples, exhibits more erratic fairness–accuracy trade-offs due to its small size and higher risk of overfitting. To better capture variations, we varied \( \lambda \) from 0 to 2.0 using the same number of points as in other datasets, effectively doubling the step size (e.g., 0.2, 0.4, 0.6, \(\ldots\)). In both outcome-based (left) and EOO-based (right) settings, the curves begin at \( \lambda = 0 \) with the typical model’s baseline accuracy (\(\approx 80\%\)) and fairness for each metric. Baseline fairness spans a relatively narrow range—about 0.57 to 0.71 in the outcome-based case, and about 0.60 to 0.73 in the EOO-based case. While fluctuations are more visible here than in the larger datasets, reflecting the impact of the smaller sample size, the overall pattern is consistent: as fairness improves with increasing \( \lambda \), accuracy tends to decline. The steepness of this decline varies across metrics and \( \lambda \) ranges, but no single metric consistently dominates in terms of minimizing accuracy loss. This variability underscores the need for careful \( \lambda \) tuning in low-data regimes, where both fairness and accuracy can be more sensitive to regularization strength.
\end{itemize}

Overall, while the trade-off between accuracy and fairness is present across all datasets, its severity and onset depend on both the fairness metric’s baseline level in the typical model and the characteristics of the dataset. This reinforces the utility of a flexible framework that enables stakeholders to explore multiple fairness–accuracy configurations before selecting an operational setting.

\begin{figure}[p]
    \centering

    \scalebox{0.82}{ 
    \begin{subfigure}[b]{\textwidth}
        \centering
        \begin{subfigure}[b]{0.45\textwidth}
            \centering
            \includegraphics[width=\textwidth]{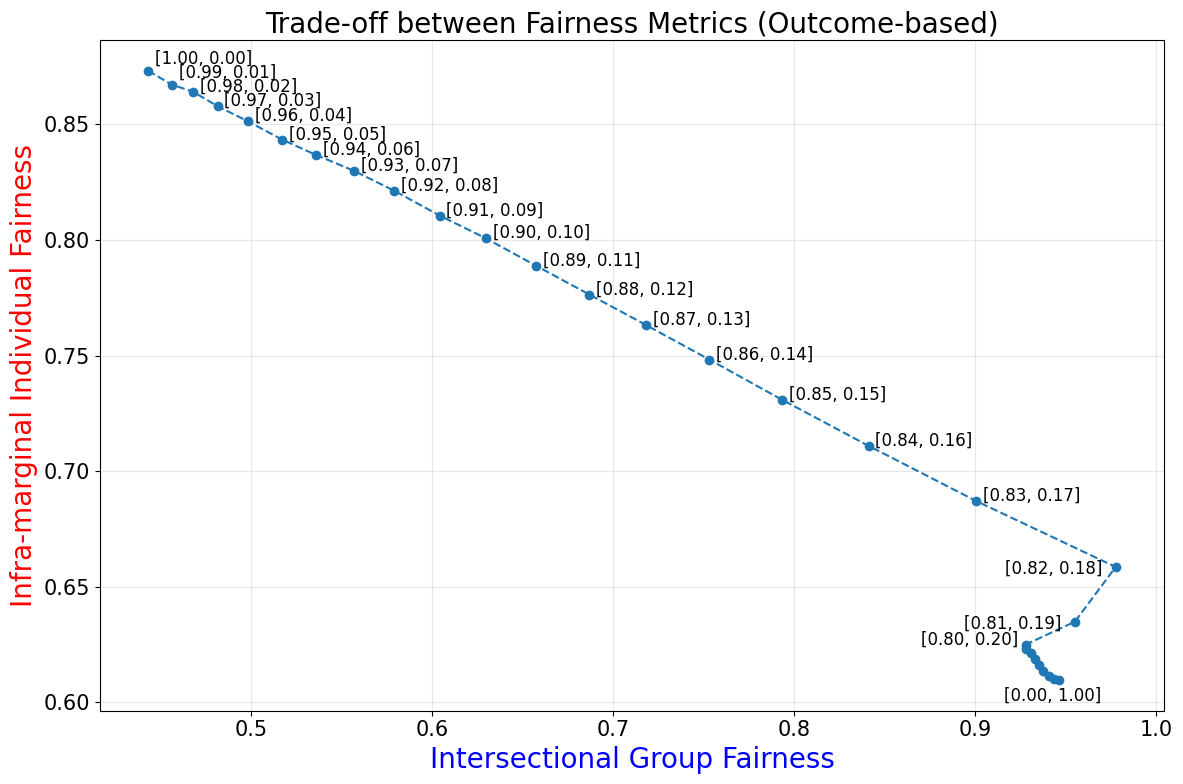}
        \end{subfigure}
        \hfill
        \begin{subfigure}[b]{0.45\textwidth}
            \centering
            \includegraphics[width=\textwidth]{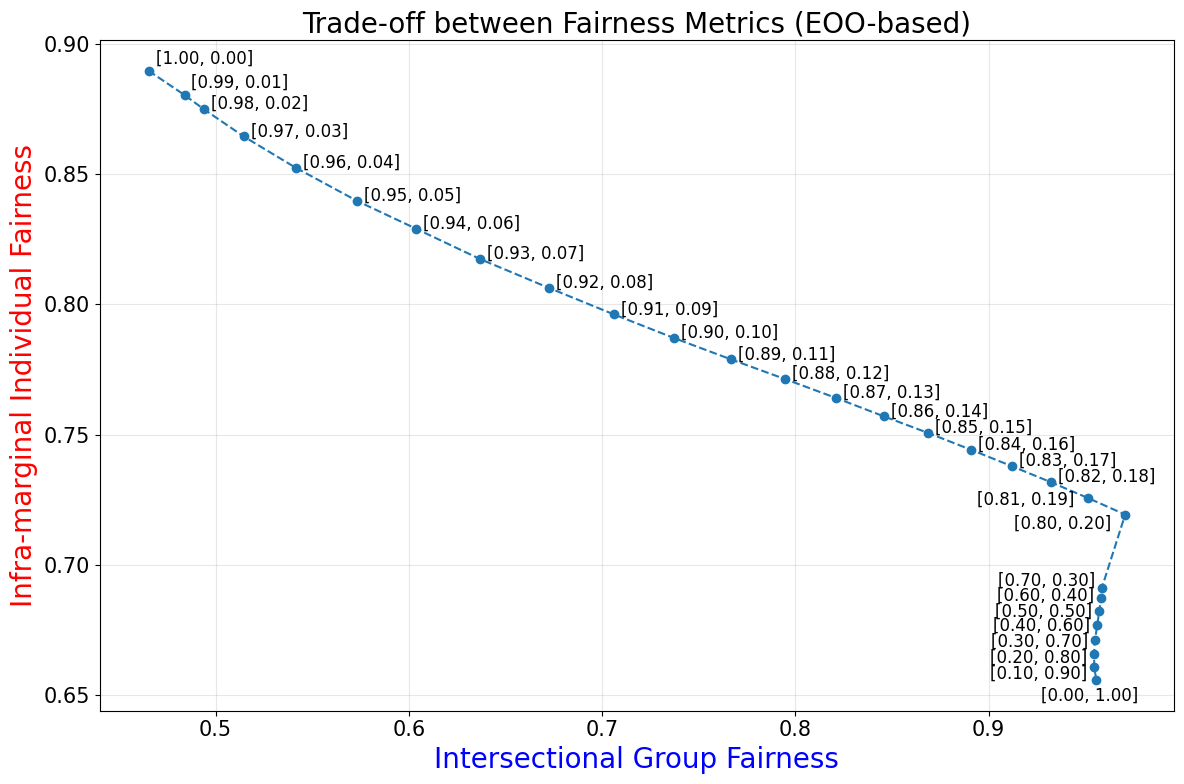}
        \end{subfigure}
        \caption{ProPublica COMPAS Dataset}
        \label{fig:compas_fairness_tradeoff}
    \end{subfigure}
    }

    \scalebox{0.82}{ 
    \begin{subfigure}[b]{\textwidth}
        \centering
        \begin{subfigure}[b]{0.45\textwidth}
            \centering
            \includegraphics[width=\textwidth]{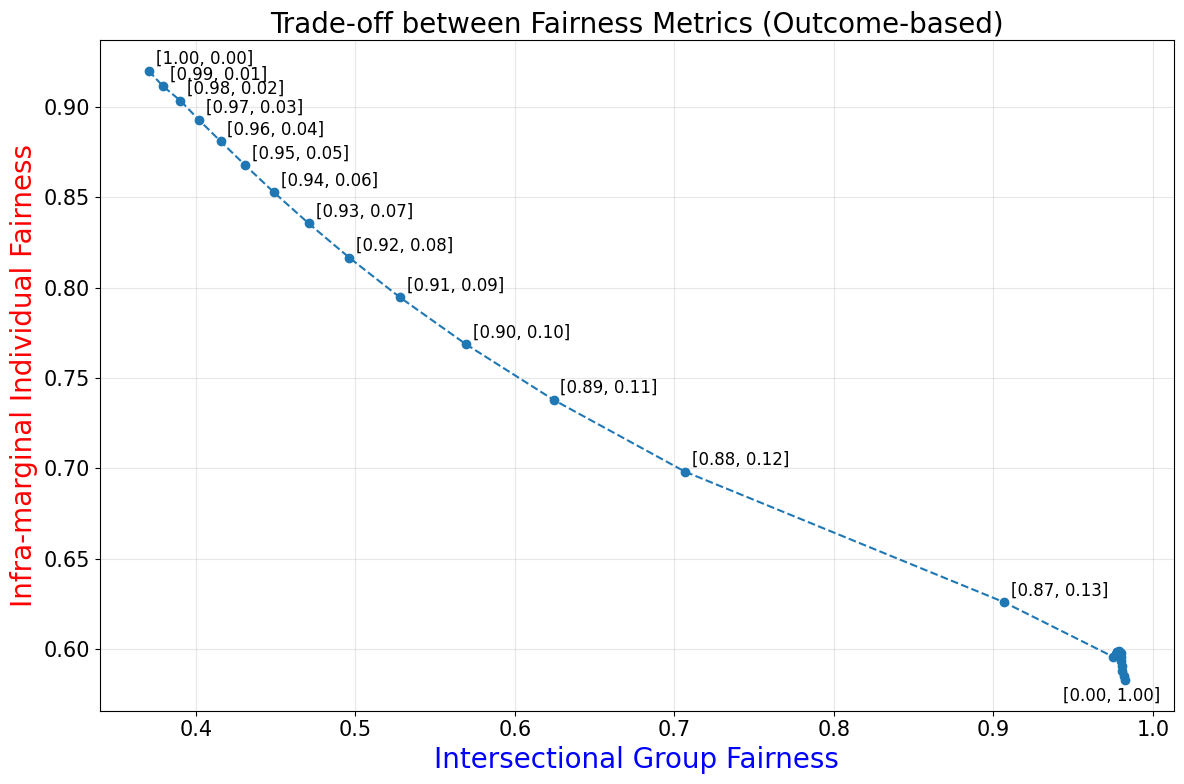}
        \end{subfigure}
        \hfill
        \begin{subfigure}[b]{0.45\textwidth}
            \centering
            \includegraphics[width=\textwidth]{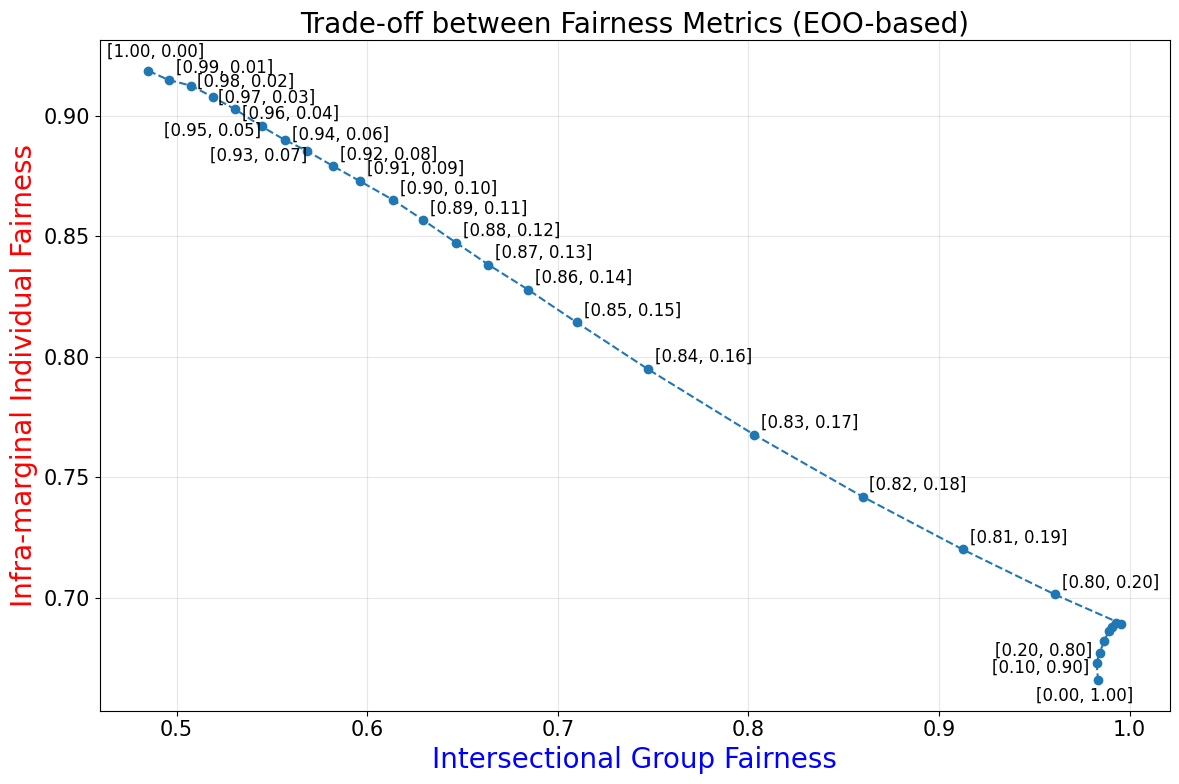}
        \end{subfigure}
        \caption{UCI Adult Dataset}
        \label{fig:adult_fairness_tradeoff}
    \end{subfigure}
    }

    \scalebox{0.82}{ 
    \begin{subfigure}[b]{\textwidth}
        \centering
        \begin{subfigure}[b]{0.45\textwidth}
            \centering
            \includegraphics[width=\textwidth]{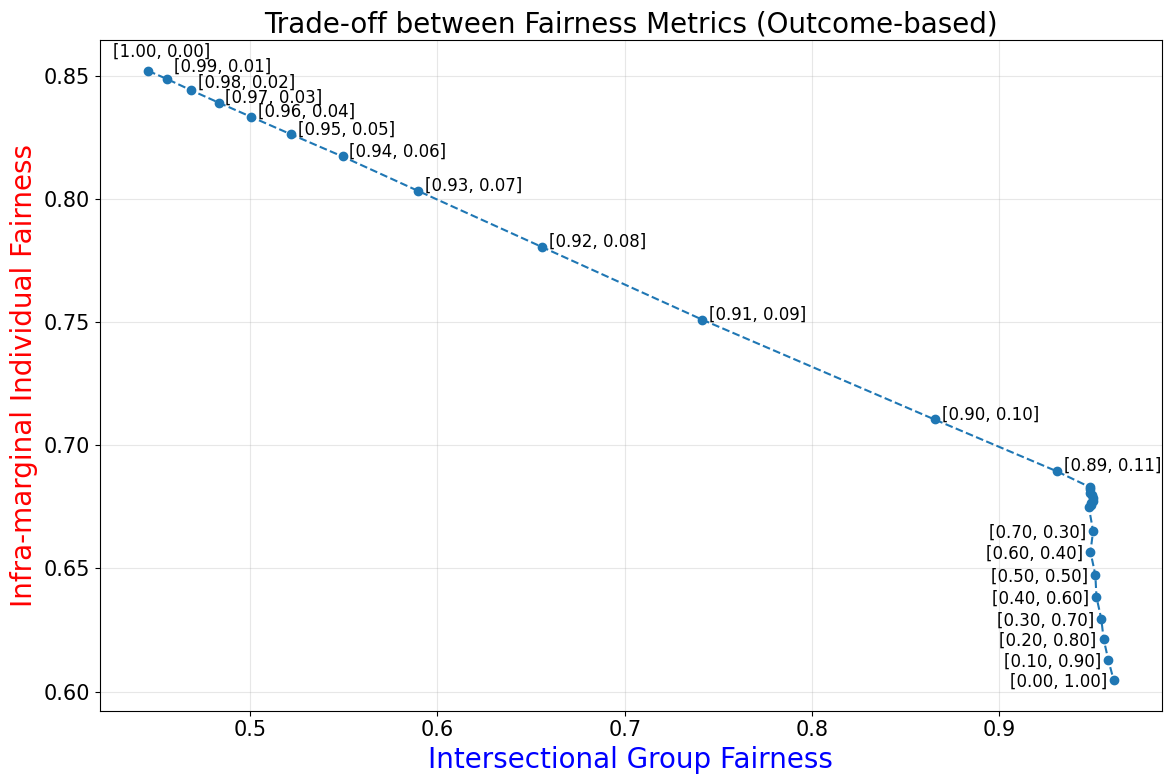}
        \end{subfigure}
        \hfill
        \begin{subfigure}[b]{0.45\textwidth}
            \centering
            \includegraphics[width=\textwidth]{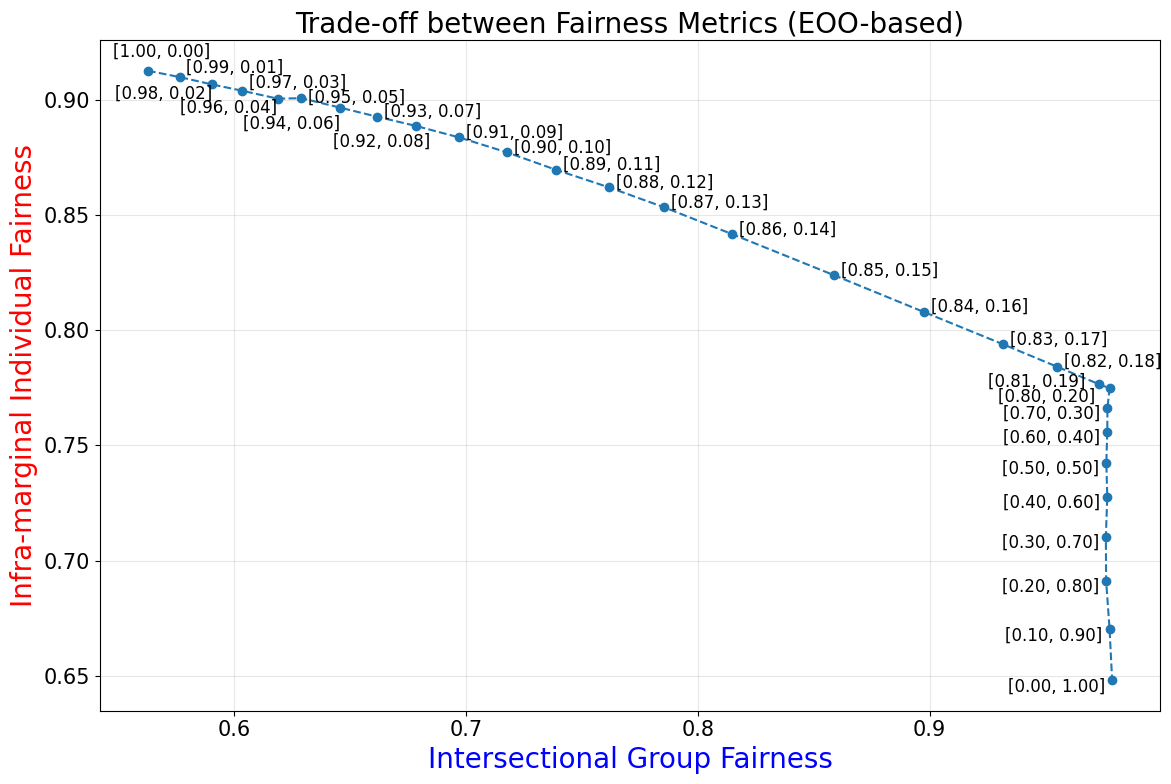}
        \end{subfigure}
        \caption{MEPS Dataset}
        \label{fig:meps_fairness_tradeoff}
    \end{subfigure}
    }

    \scalebox{0.82}{ 
    \begin{subfigure}[b]{\textwidth}
        \centering
        \begin{subfigure}[b]{0.45\textwidth}
            \centering
            \includegraphics[width=\textwidth]{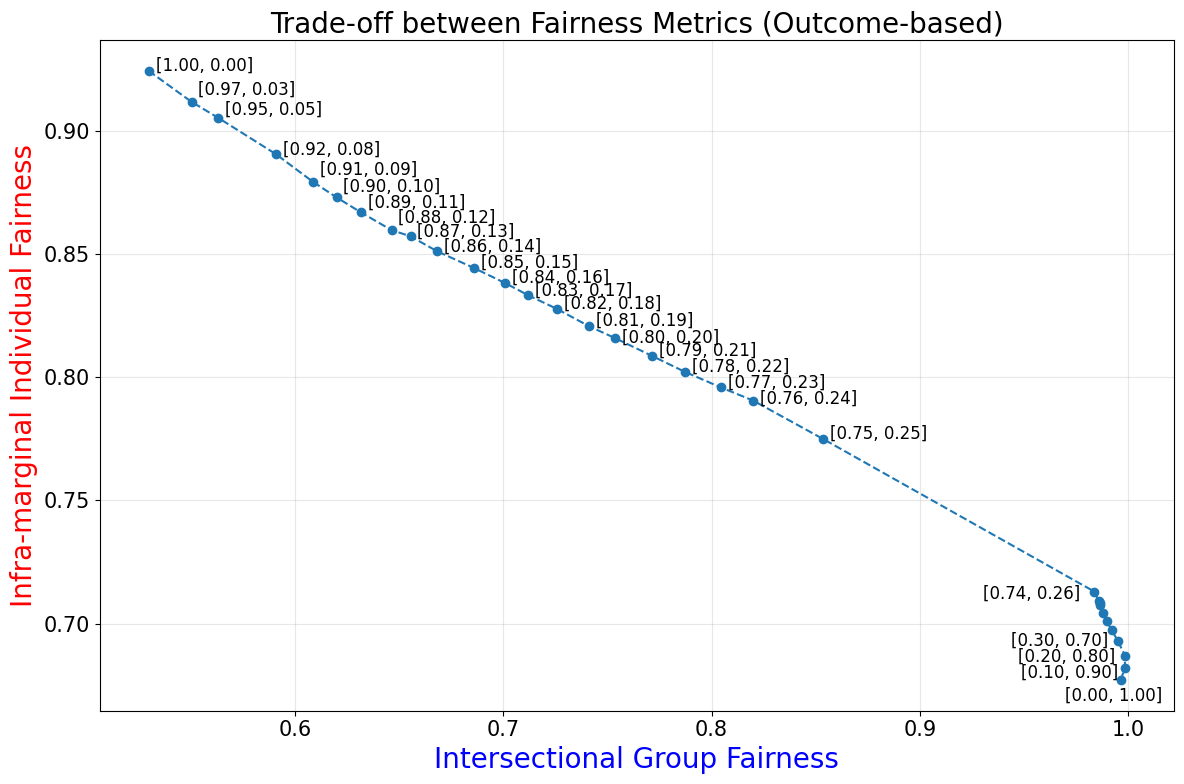}
        \end{subfigure}
        \hfill
        \begin{subfigure}[b]{0.45\textwidth}
            \centering
            \includegraphics[width=\textwidth]{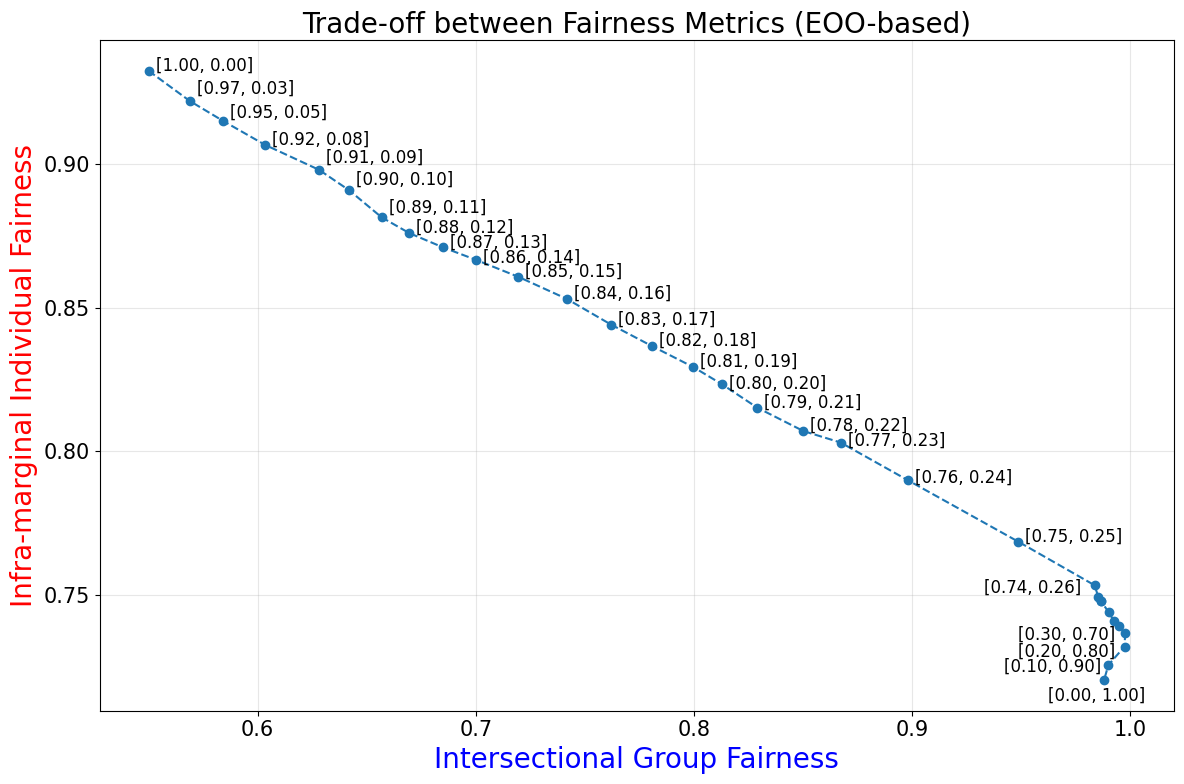}
        \end{subfigure}
        \caption{German Credit Dataset}
        \label{fig:german_fairness_tradeoff}
    \end{subfigure}
    }

    \caption{Fairness trade-offs between infra-marginal individual fairness and intersectional group fairness across four datasets. Left column: outcome-based trade-offs; right column: equality-of-opportunity-based trade-offs. Each point corresponds to a different fairness weight configuration ([$w_{\mbox{I-M,ind}},  w_{\mbox{int,ind}}, w_{\mbox{I-M,grp}}$, $w_{\mbox{int,grp}}$]) with $w_{\mbox{int,ind}}, w_{\mbox{I-M,grp}}$ kept fixed at 0.}
    \label{fig:fairness_metric_tradeoff}
\end{figure}

\begin{figure}[tb]
    \centering

    \begin{subfigure}[b]{\textwidth}
        \centering
        \begin{subfigure}[b]{0.45\textwidth}
            \centering
            \includegraphics[width=\textwidth]{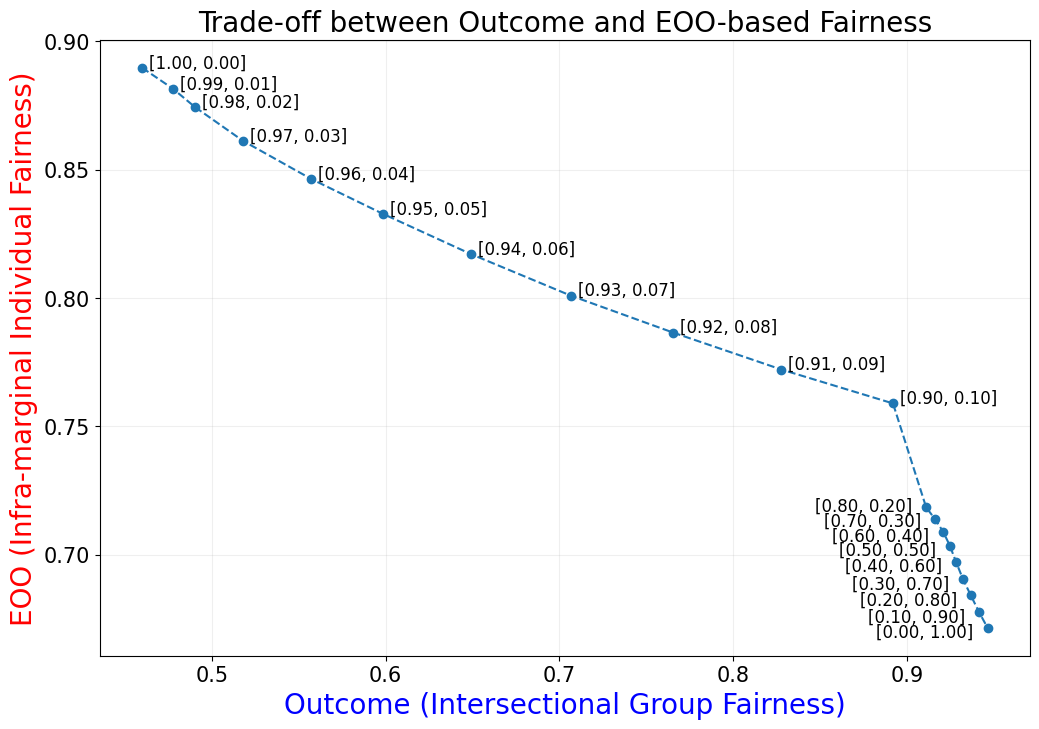}
            \caption{COMPAS Dataset}
            \label{fig:compas_outcome_vs_eoo}
        \end{subfigure}
        \hfill
        \begin{subfigure}[b]{0.45\textwidth}
            \centering
            \includegraphics[width=\textwidth]{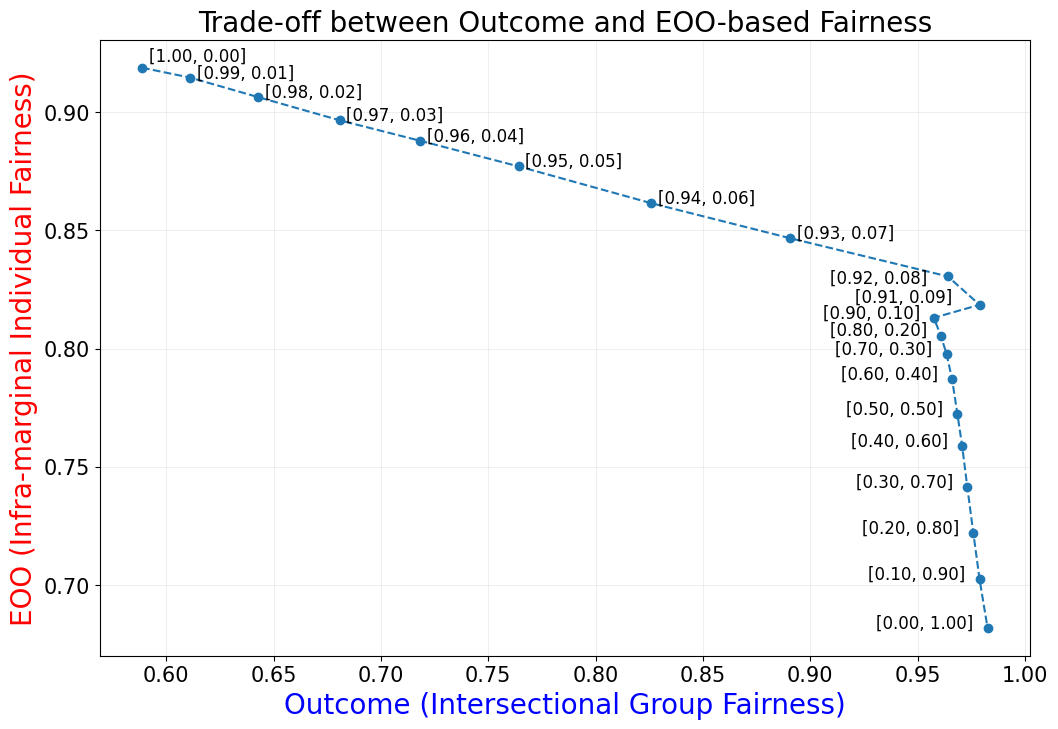}
            \caption{UCI Adult Dataset}
            \label{fig:adult_outcome_vs_eoo}
        \end{subfigure}
    \end{subfigure}

    \vspace{1em}

    \begin{subfigure}[b]{\textwidth}
        \centering
        \begin{subfigure}[b]{0.45\textwidth}
            \centering
            \includegraphics[width=\textwidth]{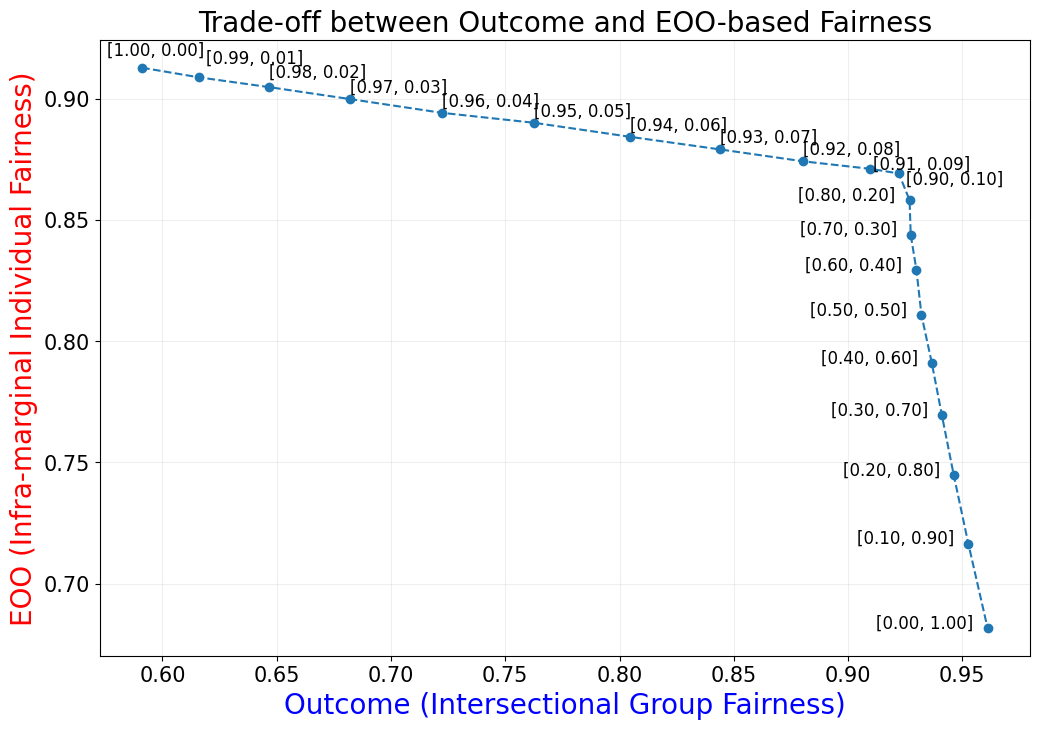}
            \caption{MEPS Dataset}
            \label{fig:meps_outcome_vs_eoo}
        \end{subfigure}
        \hfill
        \begin{subfigure}[b]{0.45\textwidth}
            \centering
            \includegraphics[width=\textwidth]{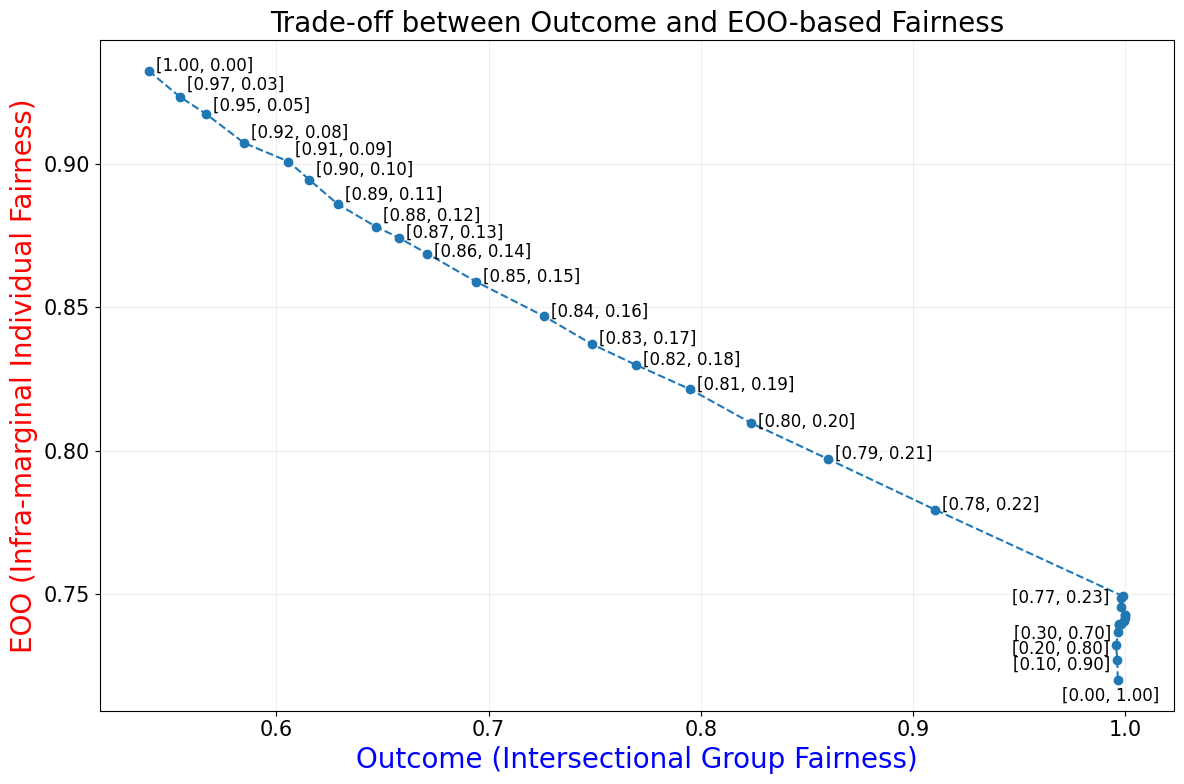}
            \caption{German Credit Dataset}
            \label{fig:german_outcome_vs_eoo}
        \end{subfigure}
    \end{subfigure}

    \caption{Fairness relationships between outcome-based and EOO-based intersectional group fairness across four datasets. Each point corresponds to a different fairness weight configuration ([$w_{\mbox{I-M,ind}},  w_{\mbox{int,ind}}, w_{\mbox{I-M,grp}}$, $w_{\mbox{int,grp}}$]) with $w_{\mbox{int,ind}}, w_{\mbox{I-M,grp}}$ kept fixed at 0.}
    \label{fig:eoo_vs_outcome_tradeoffs}
\end{figure}

\subsection{Fairness Metric Trade-Off}
We now investigate trade-offs between different fairness metrics when seeking to enforce them simultaneously, as required to achieve multi-stakeholder fairness compromises, enabled by our framework. Figure~\ref{fig:fairness_metric_tradeoff} visualizes the trade-off between two key fairness metrics: infra-marginal individual fairness (y-axis) and intersectional group fairness (x-axis), under varying weight configurations. As described in Equation~\ref{eq:pairwise_fairness}, we fix the fairness regularization strength \( \lambda \) and sweep the trade-off parameter \( \alpha \), such that \( w_a = \alpha \), \( w_b = 1 - \alpha \), and the remaining weights are zero.

This experiment illustrates a core (though expected) insight: improving one fairness metric often comes at the cost of another, underscoring inherent tensions between fairness goals. Interestingly, however, in a few cases including the COMPAS dataset, as the weight for intersectional group fairness approached 1, increasing its weight further was detrimental to both fairness metrics under consideration. In other words, when seeking to improve one metric, in some cases it was beneficial to give a non-zero weight to a different metric. We plan to investigate this phenomenon further in future work. Using our framework, these curves can help stakeholders to identify configurations that strike an efficient balance, ideally one that is an acceptable compromise for multiple stakeholders with competing fairness goals. While this figure focuses on a specific pair of metrics, similar analyses can be conducted for other combinations.

\section{Case Study: Exploring Fairness Implications in Datasets}

To demonstrate the practical implications of our framework, we present two case studies from high-stakes domains: criminal justice (COMPAS) and healthcare (MEPS). Each case examines how different stakeholders tend to prioritize distinct fairness metrics and how our weighted optimization accommodates and balances these preferences. For clarity, we model each stakeholder with a single representative metric; this simplifies real-world diversity of views. These mappings are hypothetical but literature-informed choices based on our reading of the cited sources. The framework can also encode \emph{multi-metric} preferences for a single stakeholder.

We conducted a grid search over normalized weight vectors $\mathbf{w}$ (nonnegative, $\sum_i w_i=1$) and a small set of $\lambda$ values. For each stakeholder, we selected the configuration that maximized the stakeholder’s target fairness metric subject to a pre-specified accuracy tolerance relative to a no-fairness baseline (tie-break: highest accuracy). We report held-out performance for the chosen settings. For each stakeholder, we select on a held-out validation set the configuration that maximizes the stakeholder’s target fairness metric while keeping accuracy within 5 percentage points of the $\lambda{=}0$ (no-fairness) baseline; we then report performance on the test set.

\subsection{COMPAS Recidivism Predictions and Fairness Trade-offs}

Our first case study examines the COMPAS recidivism dataset, a widely studied benchmark for fairness in criminal justice. This dataset is used to predict whether a defendant is likely to re-offend, and has been at the center of public debates around algorithmic bias. Within this context, stakeholders such as public safety advocates, civil rights organizations, and social workers often prioritize competing fairness criteria based on their respective institutional goals and ethical commitments. To reflect these variations, we associate each stakeholder with a distinct fairness metric, spanning the spectrum of individual vs. group, infra-marginal vs. intersectional, and outcome-based vs. equality-of-opportunity (EOO)-based definitions. In this case study, each stakeholder is matched to a single fairness metric that best aligns with their values.

\vspace{0.8em}
\noindent \textbf{Stakeholders and Fairness Preferences:}
\begin{itemize}
    \item \textbf{Public Safety Advocates} aim to minimize crime by accurately detaining high-risk individuals. They generally emphasize consistent, merit-based decision-making—ensuring that similarly risky individuals receive equivalent treatment regardless of their demographics. This aligns with \textbf{EOO-based infra-marginal individual fairness}, emphasizing parity in treatment among true positives across demographic groups, without introducing adjustments for historical bias. \cite{berk2021fairness}. This fairness metric matches individuals based on “fair” features (e.g., prior offenses) and evaluates fairness only on true positives. It reflects a conservative perspective that assumes a level playing field and prioritizes minimizing prediction errors among similarly qualified individuals \cite{barocas-hardt-narayanan,corbett2023measure}.
    
    \item \textbf{Civil Rights Organizations} often seek to mitigate systemic discrimination and ensure that marginalized groups are not disproportionately penalized. They prioritize \textbf{outcome-based intersectional group fairness}, which seeks parity in outcomes across demographic groups, regardless of underlying base rates or the correctness of predictions. This fairness notion reflects a commitment to correcting historical inequities and aligns with affirmative action perspectives frequently endorsed by civil rights and social justice advocates \cite{hardt2016equality,barocas2020hidden,noble2018algorithms,mitchell2021algorithmic,aclu2020bias}.
    
    \item \textbf{Social Workers or Rehabilitation Advocates} often emphasize equitable access to support services such as job training, mental health care, or housing. They advocate for fairness in how post-release interventions are distributed among those at actual risk of recidivism. This aligns with \textbf{EOO-based intersectional group fairness}, which ensures that across demographic lines, individuals who truly re-offend receive equal access to rehabilitative resources. This fairness metric does not rely on individual feature matching but instead averages outcomes across intersectional subgroups, emphasizing demographic parity in true positive outcomes \cite{braveman2014health,obermeyer2019dissecting}.

\end{itemize}

\vspace{1em}
\subsubsection{Stakeholder-Aligned Fairness Configurations}

Figure~\ref{fig:compas_fairness_tradeoff} visualizes trade-offs between different fairness notions across weight configurations. Below, we describe how each stakeholder’s preferred fairness definition can be implemented using our framework, supported by fairness and accuracy outcomes. We first report the solutions that each stakeholder would find if they are allowed to select the trade-off hyper-parameters unilaterally, maximizing their chosen fairness metric subject to an accuracy tolerance relative to that attained by the no-fairness baseline. The selection satisfies a \(\leq\!5\) percentage-point accuracy tolerance relative to the $\lambda{=}0$ baseline.

\vspace{0.5em}
\noindent\textbf{Public Safety Advocates (EOO-Based, Infra-marginal Individual Fairness)}  
\begin{itemize}
  \item $\lambda = 1.0$, $w_{\mbox{I-M,ind}} = 1.0$, all other $w = 0$
  \item Fairness metrics: $R_{I-M,ind}^{\text{EOO}} = \textbf{0.83}$
  \item Accuracy: \textbf{94.25\%}
\end{itemize}

\vspace{0.5em}
\noindent\textbf{Civil Rights Organizations (Outcome-Based, Intersectional Group Fairness)}  
\begin{itemize}
  \item $\lambda = 3.0$, $w_{\mbox{int,grp}} = 1.0$, all other $w = 0$
  \item Fairness metrics: $R_{int,grp}^{\text{outcome}} = \textbf{0.95}$
  \item Accuracy: \textbf{96.1\%}
\end{itemize}

\vspace{0.5em}
\noindent\textbf{Social Workers / Rehabilitation Advocates (EOO-Based, Intersectional Group Fairness)}  
\begin{itemize}
  \item $\lambda = 3.0$, $w_{\mbox{int,grp}} = 1.0$, all other $w = 0$
  \item Fairness metrics: $R_{int,grp}^{\text{EOO}} = \textbf{0.96}$
  \item Accuracy: \textbf{92.96\%}
\end{itemize}

\vspace{1em}
\subsubsection{Consensus Solution: Balancing Stakeholder Fairness Preferences}

In practice, competing stakeholder interests may require balancing fairness notions. For example, Public Safety Advocates typically emphasize \textbf{EOO-based infra-marginal individual fairness}, ensuring that among true positives, similarly risky individuals are treated consistently. In contrast, Civil Rights Organizations prioritize \textbf{outcome-based intersectional group fairness}, which aims to equalize outcomes across demographic groups regardless of ground truth.

To explore this tension, and how multi-stakeholder compromise solutions might be attained, we conducted an experiment varying the weights assigned to these two fairness metrics while fixing $\lambda = 3.0$ for the consensus trade-off analysis. Note that in the standalone stakeholder results above, Public Safety Advocates were represented at $\lambda = 1.0$, whereas here we use $\lambda = 3.0$ for consistency when comparing against Civil Rights Organizations and when evaluating balanced configurations. Specifically, we set $w_{\mbox{I-M,ind}} + w_{\mbox{int,grp}} = 1.0$ and sweep their trade-off from 0.9–0.1 to 0.1–0.9.

Key configurations from this trade-off exploration, which might constitute chosen solutions, are as follows:

\begin{itemize}
  \item \textbf{Public Safety–Dominant:}  
  $w_{\mbox{I-M,ind}} = 0.9$, $w_{\mbox{int,grp}} = 0.1$:  
  $R_{I-M,ind}^{\text{EOO}} = \textbf{0.76}$, $R_{int,grp}^{\text{outcome}} = \textbf{0.89}$, Accuracy = \textbf{92.62\%}

  \item \textbf{Balanced Configuration:}  
  $w_{\mbox{I-M,ind}} = 0.5$, $w_{\mbox{int,grp}} = 0.5$:  
  $R_{I-M,ind}^{\text{EOO}} = \textbf{0.70}$, $R_{int,grp}^{\text{outcome}} = \textbf{0.92}$, Accuracy = \textbf{95.66\%}

  \item \textbf{Civil Rights–Dominant:}  
  $w_{\mbox{I-M,ind}} = 0.1$, $w_{\mbox{int,grp}} = 0.9$:  
  $R_{I-M,ind}^{\text{EOO}} = \textbf{0.68}$, $R_{int,grp}^{\text{outcome}} = \textbf{0.94}$, Accuracy = \textbf{96.31\%}
\end{itemize}

These results illustrate how the framework enables both stakeholder-specific configurations and cross-stakeholder compromise. Public Safety Advocates achieve their preferred EOO-based infra-marginal individual fairness at $\lambda = 1.0$, yielding strong accuracy (94.25\%) with moderate fairness ($R_{I\text{-}M,ind}^{\text{EOO}} = 0.83$). In contrast, Civil Rights Organizations and Social Workers emphasize intersectional group fairness under $\lambda = 3.0$, where fairness scores are high (0.95--0.96) while accuracy remains competitive (92.96--96.10\%). When fairness notions are combined in the consensus experiment at $\lambda = 3.0$, the trade-off curve shows that shifting weights between individual- and group-based criteria moves performance smoothly along a spectrum. This flexibility enables stakeholders to emphasize what matters most to them—be it consistency among individuals, parity across groups, or predictive performance. By exploring the trade-offs, multiple stakeholders may seek a compromise solution.

\subsection{High Medical Utilization Predictions and Fairness Trade-offs (MEPS)}

Our second case study focuses on the Medical Expenditure Panel Survey (MEPS) dataset, which provides a rich view of healthcare access and usage across U.S. populations. We predict whether an individual will require high medical utilization, defined as 10 or more medical visits in a year—a task with significant implications for resource planning and equity in care. In this healthcare setting, stakeholders such as healthcare providers, public health officials, and patient advocacy groups may prioritize different fairness goals depending on their mission. As in the previous case, we associate each stakeholder with a fairness metric that reflects their perspective, allowing us to examine how the framework facilitates negotiation and compromise in a complex policy environment.

\vspace{0.8em}
\noindent \textbf{Stakeholders and Fairness Preferences:}
\begin{itemize}
    \item \textbf{Healthcare Providers} aim to consistently identify patients with similar health conditions who are likely to require frequent care. This supports early intervention strategies such as chronic disease management. Their goals align with \textbf{EOO-based infra-marginal individual fairness}, which ensures that true positives with similar needs receive comparable predictions \cite{rajkomar2018ensuring}.
    
    \item \textbf{Public Health Officials} prioritize reducing disparities in access to care. They favor \textbf{outcome-based intersectional group fairness}, aiming to ensure that disadvantaged demographic subgroups are not under-identified, even at some cost to predictive accuracy  \cite{braveman2014health,veinot2018good,char2018implementing}.

    \item \textbf{Patient Advocacy Groups} emphasize equitable access to follow-up services such as financial aid, telehealth, or home care. They support \textbf{EOO-based intersectional group fairness}, ensuring that among those who truly need care, marginalized subgroups are not overlooked \cite{obermeyer2019dissecting,mitchell2021algorithmic}.
\end{itemize}

\vspace{1em}
\subsubsection{Stakeholder-Aligned Fairness Configurations}

Figure~\ref{fig:meps_fairness_tradeoff} visualizes trade-offs between fairness metrics for various weight settings. Below, we report fairness and accuracy outcomes for stakeholder-aligned configurations, which are selected by optimizing the stakeholder's chosen fairness metrics, subject to accuracy constraints. The selection satisfies a \(\leq\!5\) percentage-point accuracy tolerance relative to the $\lambda{=}0$ baseline.

\vspace{0.5em}
\noindent\textbf{Healthcare Providers (EOO-Based, Infra-marginal Individual Fairness)}  
\begin{itemize}
  \item $\lambda = 2.0$, $w_{\mbox{I-M,ind}} = 1.0$, all other $w = 0$
  \item Fairness metrics: $R_{I-M,ind}^{\text{EOO}} = \textbf{0.91}$
  \item Accuracy: \textbf{82.26\%}
\end{itemize}

\vspace{0.5em}
\noindent\textbf{Public Health Officials (Outcome-Based, Intersectional Group Fairness)}  
\begin{itemize}
  \item $\lambda = 2.0$, $w_{\mbox{int,grp}} = 1.0$, all other $w = 0$
  \item Fairness metrics: $R_{int,grp}^{\text{outcome}} = \textbf{0.96}$
  \item Accuracy: \textbf{84.56\%}
\end{itemize}

\vspace{0.5em}
\noindent\textbf{Patient Advocacy Groups (EOO-Based, Intersectional Group Fairness)}  
\begin{itemize}
  \item $\lambda = 2.0$, $w_{\mbox{int,grp}} = 1.0$, all other $w = 0$
  \item Fairness metrics: $R_{int,grp}^{\text{EOO}} = \textbf{0.98}$
  \item Accuracy: \textbf{86\%}
\end{itemize}

\vspace{1em}
\subsubsection{Consensus Solution: Balancing Stakeholder Fairness Preferences}

When competing values must be reconciled, a balanced configuration can help stakeholders reach consensus. In the MEPS scenario, Healthcare Providers prefer \textbf{EOO-based infra-marginal individual fairness}, while Public Health Officials advocate for \textbf{outcome-based intersectional group fairness}. We analyze their trade-off by fixing $\lambda = 2.0$ and varying $w_{\mbox{I-M,ind}} + w_{\mbox{int,grp}} = 1.0$.\\

Representative configurations include:

\begin{itemize}
  \item \textbf{Healthcare Provider–Dominant:}  
  $w_{\mbox{I-M,ind}} = 0.9$, $w_{\mbox{int,grp}} = 0.1$:  
  $R_{I-M,ind}^{\text{EOO}} = \textbf{0.87}$, $R_{int,grp}^{\text{outcome}} = \textbf{0.92}$, Accuracy = \textbf{84.69\%}

  \item \textbf{Balanced Configuration:}  
  $w_{\mbox{I-M,ind}} = 0.5$, $w_{\mbox{int,grp}} = 0.5$:  
  $R_{I-M,ind}^{\text{EOO}} = \textbf{0.81}$, $R_{int,grp}^{\text{outcome}} = \textbf{0.93}$, Accuracy = \textbf{85.07\%}

  \item \textbf{Public Health–Dominant:}  
  $w_{\mbox{I-M,ind}} = 0.1$, $w_{\mbox{int,grp}} = 0.9$:  
  $R_{I-M,ind}^{\text{EOO}} = \textbf{0.72}$, $R_{int,grp}^{\text{outcome}} = \textbf{0.95}$, Accuracy = \textbf{85.41\%}
\end{itemize}

In MEPS, adjusting the weight vector yields different fairness–accuracy trade-offs, but accuracy and fairness remain high across stakeholder configurations. 
This makes the trade-off space less contentious, giving stakeholders more freedom to emphasize their preferred fairness objective without major sacrifices in performance. 
Groups may still differ in their choices—whether they emphasize parity across groups, consistency among similarly qualified individuals, or predictive performance—but the underlying compromises are comparatively mild.

\section{Conclusion}
We introduced a unifying, \emph{human-centered} fairness framework that (i) provides a consistent, ratio-based foundation (in the spirit of the $p\%$/80\% rule) from which all eight metrics arise as combinations of three \emph{dimensions}—individual vs.\ group, infra-marginal vs.\ intersectional, and outcome-based vs.\ equality-of-opportunity (EOO); (ii) simplifies the learning curve for non-experts via shared notation and interpretable, comparable visualizations; (iii) maps metrics to human values by making normative choices explicit at the level of these dimensions; and (iv) supports multi-stakeholder compromise through a weighted, multi-objective learning setup that makes trade-offs transparent rather than implicit.

Through stakeholder-grounded case studies, we illustrated how different groups may prioritize distinct objectives and how our method can help stakeholders navigate these tensions by varying metric weights and the overall fairness hyperparameter $\lambda$. To select configurations, we used a transparent search procedure (grid search over weights with a pre-specified accuracy tolerance). (We qualify that the stakeholder preferences we identified are our best literature-informed approximations rather than definitive claims about any one group.) The fairness–fairness and fairness–accuracy curves reveal Pareto-like frontiers and, in several settings, metric complementarity: introducing a second (or third) fairness criterion can reinforce a primary objective, improving it with minimal impact on others. These tools give practitioners a concrete way to balance equity and utility in high-stakes domains.

Our findings highlight that fairness is not a fixed target but a contextual, value-sensitive negotiation. As a result, algorithmic fairness cannot be divorced from stakeholder engagement. By explicitly encoding stakeholder preferences and enabling principled compromises, our framework contributes to building more accountable and inclusive AI systems.

In future work, we plan to (1) evaluate the framework in prospective deployments with monitoring and governance; (2) systematically probe metric complementarity via grid search and multi-metric weighting to characterize regimes of reinforcement versus sharp trade-offs; and (3) conduct user studies to assess how stakeholders interpret and act on the trade-off visualizations, and refine the decision-support interface accordingly.

\section*{Acknowledgments}
This material is based upon work supported by the National Science Foundation under Grant No.’s IS1927486; IIS2046381. Any opinions, findings, and conclusions or recommendations expressed in this material are those of the author(s) and do not necessarily reflect the views of the National Science Foundation.


\bibliographystyle{plainnat}
\bibliography{main}

@String{Computing = "Computing" }

@String{Computer = "{IEEE} Computer" }

@String{Springer = "Springer-Verlag" }

@inproceedings{holstein2019improving,
  author    = {Kenneth Holstein and Jennifer Wortman Vaughan and Hal Daum{\'e} III and Miroslav Dud{\'\i}k and Hanna Wallach},
  title     = {{Improving Fairness in Machine Learning Systems: What Do Industry Practitioners Need?}},
  booktitle = {Proceedings of the 2019 {ACM} {CHI} Conference on Human Factors in Computing Systems (CHI '19)},
  year      = {2019},
  publisher = {{ACM}},
  address   = {New York, NY, USA},
  doi       = {10.1145/3290605.3300830}
}

@inproceedings{mitchell2019model,
  title={Model cards for model reporting},
  author={Mitchell, Margaret and Wu, Simone and Zaldivar, Andrew and Barnes, Parker and Vasserman, Lucy and Hutchinson, Ben and Spitzer, Elena and Raji, Inioluwa Deborah and Gebru, Timnit},
  booktitle={Proceedings of the conference on fairness, accountability, and transparency},
  pages={220--229},
  year={2019},
  publisher = {{ACM}}
}

@article{gebru2021datasheets,
  author    = {Timnit Gebru and Jamie Morgenstern and Briana Vecchione and Jennifer Wortman Vaughan and Hanna Wallach and Hal Daum{\'e} III and Kate Crawford},
  title     = {Datasheets for Datasets},
  journal   = {Communications of the {ACM}},
  year      = {2021},
  volume    = {64},
  number    = {12},
  pages     = {86--92},
  doi       = {10.1145/3458723}
}

@inproceedings{selbst2019fairness,
  title={Fairness and abstraction in sociotechnical systems},
  author={Selbst, Andrew D and Boyd, Danah and Friedler, Sorelle A and Venkatasubramanian, Suresh and Vertesi, Janet},
  booktitle={Proceedings of the conference on fairness, accountability, and transparency},
  pages={59--68},
  year={2019}
}

@inproceedings{madaio2020co,
  title     = {Co-designing Checklists to Understand Organizational Challenges and Opportunities Around Fairness in {AI}},
  author    = {Madaio, Michael A and Stark, Luke and Vaughan, Jennifer Wortman and Wallach, Hanna},
  booktitle = {Proceedings of the 2020 {ACM} {CHI} Conference on Human Factors in Computing Systems ({CHI} '20)},
  pages     = {1--14},
  year      = {2020},
  publisher = {{ACM}}
}

@inproceedings{feldman2015certifying,
  title     = {Certifying and Removing Disparate Impact},
  author    = {Feldman, Michael and Friedler, Sorelle A. and Moeller, John and Scheidegger, Carlos and Venkatasubramanian, Suresh},
  booktitle = {Proceedings of the 21st {ACM} {SIGKDD} International Conference on Knowledge Discovery and Data Mining ({KDD} '15)},
  pages     = {259--268},
  year      = {2015},
  publisher = {{ACM}}
}

@article{calders2010three,
  title={Three naive bayes approaches for discrimination-free classification},
  author={Calders, Toon and Verwer, Sicco},
  journal={Data mining and knowledge discovery},
  volume={21},
  number={2},
  pages={277--292},
  year={2010},
  publisher={Springer}
}

@inproceedings{zafar2017fairnessbeyond,
  author    = {Muhammad Bilal Zafar and Isabel Valera and Manuel Gomez Rodriguez and Krishna P. Gummadi},
  title     = {{Fairness Beyond Disparate Treatment \& Disparate Impact: Learning Classification without Disparate Mistreatment}},
  booktitle = {Proceedings of the 26th International World Wide Web Conference ({WWW} '17)},
  year      = {2017},
  pages     = {1171--1180},
  publisher = {ACM},
  address   = {Perth, Australia},
  doi       = {10.1145/3038912.3052660}
}

@article{lee2019webuildai,
  author    = {Min Kyung Lee and Daniel Kusbit and Anson Kahng and Ji Tae Kim and Xinran Yuan and Allissa Chan and Daniel See and Ritesh Noothigattu and Siheon Lee and Alexandros Psomas and Ariel D. Procaccia},
  title     = {{WeBuildAI: Participatory Framework for Algorithmic Governance}},
  journal   = {Proceedings of the ACM on Human-Computer Interaction},
  year      = {2019},
  volume    = {3},
  number    = {CSCW},
  articleno = {181},
  numpages  = {35},
  publisher = {{ACM}},
  doi       = {10.1145/3359283}
}

@inproceedings{rahman2024towards,
  title={Towards A Unifying Human-Centered AI Fairness Framework},
  author={Rahman, Munshi Mahbubur and Pan, Shimei and Foulds, James R},
  booktitle={Proceedings of the 2024 International Conference on Information Technology for Social Good},
  pages={88--92},
  year={2024}
}

@techreport{ainow2023landscape,
  title        = {AI Now 2023 Landscape Report},
  author       = {{AI Now Institute}},
  year         = {2023},
  institution  = {AI Now Institute},
  url          = {https://ainowinstitute.org/wp-content/uploads/2023/04/AI-Now-2023-Landscape-Report-FINAL.pdf}
}

@misc{statlog_german_credit_data_144,
  author       = {Hofmann, Hans},
  title        = {{Statlog (German Credit Data)}},
  year         = {1994},
  howpublished = {UCI Machine Learning Repository},
  note         = {{DOI}: https://doi.org/10.24432/C5NC77}
}

@inproceedings{bellamy2018ai,
  title     = {{AI Fairness 360: An Extensible Toolkit for Detecting, Understanding, and Mitigating Unwanted Algorithmic Bias}},
  author    = {Bellamy, Rachel K. E. and Dey, Kuntal and Hind, Michael and Hoffman, Samuel and Houde, Stephanie and Kannan, Suresh and Lohia, Preetam and Martino, Jacquelyn and Mehta, Sameep and Mojsilovi{\'c}, Aleksandra and others},
  booktitle = {Proceedings of the {IEEE} International Conference on Data Mining Workshops ({ICDMW})},
  pages     = {121--126},
  year      = {2018},
  organization = {{IEEE}}
}

@article{braveman2014health,
  title={Health disparities and health equity: the issue is justice},
  author={Braveman, Paula and Gruskin, Sofia},
  journal={American Journal of Public Health},
  volume={101},
  number={S1},
  pages={S149--S155},
  year={2014},
  publisher={American Public Health Association}
}

@inproceedings{hardt2016equality,
  author    = {Hardt, Moritz and Price, Eric and Srebro, Nati},
  title     = {Equality of Opportunity in Supervised Learning},
  booktitle = {Advances in Neural Information Processing Systems ({NeurIPS} 29)},
  year      = {2016}
}

@misc{narayanan2018translation,
  title={Translation tutorial: 21 fairness definitions and their politics},
  author={Narayanan, Arvind},
  year={2018},
  howpublished={Tutorial presented at Fairness, Accountability, and Transparency (FAT* 2018)},
  note={Available at \url{https://fairnesstutorial.org}}
}

@article{corbett2023measure,
  title={{The measure and mismeasure of fairness: A critical review of fair machine learning}},
  author={Corbett-Davies, Sam and Goel, Sharad},
  journal={Communications of the ACM},
  volume={66},
  number={3},
  pages={50--59},
  year={2023},
  publisher={ACM}
}

@article{barocas2020hidden,
  title={Hidden assumptions in fair determination},
  author={Barocas, Solon and Selbst, Andrew D. and Raghavan, Manish},
  journal={Harvard Journal of Law \& Technology},
  volume={33},
  number={2},
  year={2020}
}

@book{noble2018algorithms,
  title={Algorithms of oppression: How search engines reinforce racism},
  author={Noble, Safiya Umoja},
  year={2018},
  publisher={NYU Press}
}

@misc{aclu2020bias,
  title={Bias in algorithms: A troubling blind spot},
  author={ACLU},
  year={2020},
  note={Available at: \url{https://www.aclu.org/issues/privacy-technology/surveillance-technologies/bias-algorithms}}
}

@inproceedings{la2023optimizing,
  title={Optimizing fairness tradeoffs in machine learning with multiobjective meta-models},
  author={La Cava, William G},
  booktitle={Proceedings of the Genetic and Evolutionary Computation Conference},
  pages={511--519},
  year={2023}
}

@article{liu2022accuracy,
  title={{Accuracy and fairness trade-offs in machine learning: A stochastic multi-objective approach}},
  author={Liu, Suyun and Vicente, Luis Nunes},
  journal={Computational Management Science},
  volume={19},
  number={3},
  pages={513--537},
  year={2022},
  publisher={Springer}
}

@article{obermeyer2019dissecting,
  title={Dissecting racial bias in an algorithm used to manage the health of populations},
  author={Obermeyer, Ziad and Powers, Brian and Vogeli, Christine and Mullainathan, Sendhil},
  journal={Science},
  volume={366},
  number={6464},
  pages={447--453},
  year={2019},
  publisher={American Association for the Advancement of Science}
}

@inproceedings{dwork2012fairness,
  title={Fairness through awareness},
  author={Dwork, Cynthia and Hardt, Moritz and Pitassi, Toniann and Reingold, Omer and Zemel, Richard},
  booktitle={Proceedings of the 3rd innovations in theoretical computer science conference},
  pages={214--226},
  year={2012}
}

@article{berk2021fairness,
  title={{Fairness in criminal justice risk assessments: The state of the art}},
  author={Berk, Richard and Heidari, Hoda and Jabbari, Shahin and Kearns, Michael and Roth, Aaron},
  journal={Sociological Methods \& Research},
  volume={50},
  number={1},
  pages={3--44},
  year={2021},
  publisher={Sage Publications Sage CA: Los Angeles, CA}
}

@incollection{crenshaw2013demarginalizing,
  title={{Demarginalizing the intersection of race and sex: A black feminist critique of antidiscrimination doctrine, feminist theory and antiracist politics}},
  author={Crenshaw, Kimberl{\'e}},
  booktitle={Feminist legal theories},
  pages={23--51},
  year={2013},
  publisher={Routledge}
}

@book{collins2022black,
  title={Black feminist thought: Knowledge, consciousness, and the politics of empowerment},
  author={Collins, Patricia Hill},
  year={2022},
  publisher={routledge}
}

@inproceedings{buolamwini2018gender,
  title={{Gender shades: Intersectional accuracy disparities in commercial gender classification}},
  author={Buolamwini, Joy and Gebru, Timnit},
  booktitle={Conference on fairness, accountability and transparency},
  pages={77--91},
  year={2018},
  organization={PMLR}
}

@book{davis2011prisons,
  title={Are prisons obsolete?},
  author={Davis, Angela Y},
  year={2011},
  publisher={Seven stories press}
}

@article{hooks2014ain,
  title={{Ain't I a woman: Black women and feminism}},
  author={Hooks, Bell},
  year={2014}
}

@article{wald2003defining,
  title={Defining and redirecting a school-to-prison pipeline},
  author={Wald, Johanna and Losen, Daniel J},
  journal={New directions for youth development},
  volume={2003},
  number={99},
  pages={9--15},
  year={2003},
  publisher={Wiley Subscription Services, Inc., A Wiley Company Hoboken}
}

@misc{misc_adult_2,
  author       = {Becker,Barry and Kohavi,Ronny},
  title        = {{Adult}},
  year         = {1996},
  howpublished = {UCI Machine Learning Repository},
  note         = {{DOI}: https://doi.org/10.24432/C5XW20}
}

@article{islam2023differential,
  title={{Differential fairness: An intersectional framework for fair AI}},
  author={Islam, Rashidul and Keya, Kamrun Naher and Pan, Shimei and Sarwate, Anand D and Foulds, James R},
  journal={Entropy},
  volume={25},
  number={4},
  pages={660},
  year={2023},
  publisher={MDPI}
}

@book{barocas-hardt-narayanan,
  title = {Fairness and Machine Learning: Limitations and Opportunities},
  author = {Solon Barocas and Moritz Hardt and Arvind Narayanan},
  publisher = {MIT Press},
  year = {2023}
}

@article{mehrabi2021survey,
  title={A survey on bias and fairness in machine learning},
  author={Mehrabi, Ninareh and Morstatter, Fred and Saxena, Nripsuta and Lerman, Kristina and Galstyan, Aram},
  journal={ACM computing surveys (CSUR)},
  volume={54},
  number={6},
  pages={1--35},
  year={2021},
  publisher={ACM New York, NY, USA}
}

@article{vzliobaite2017measuring,
  title={Measuring discrimination in algorithmic decision making},
  author={{\v{Z}}liobait{\.e}, Indr{\.e}},
  journal={Data Mining and Knowledge Discovery},
  volume={31},
  number={4},
  pages={1060--1089},
  year={2017},
  publisher={Springer}
}

@inproceedings{binns2018fairness,
  title={Fairness in machine learning: Lessons from political philosophy},
  author={Binns, Reuben},
  booktitle={Conference on fairness, accountability and transparency},
  pages={149--159},
  year={2018},
  organization={PMLR}
}

@inproceedings{islam2021free,
author = {Islam, Rashidul and Pan, Shimei and Foulds, James R.},
title = {Can We Obtain Fairness For Free?},
year = {2021},
isbn = {9781450384735},
publisher = {Association for Computing Machinery},
address = {New York, NY, USA},
url = {https://doi.org/10.1145/3461702.3462614},
doi = {10.1145/3461702.3462614},
booktitle = {Proceedings of the 2021 AAAI/ACM Conference on AI, Ethics, and Society},
pages = {586–596},
numpages = {11},
location = {Virtual Event, USA},
series = {AIES '21}
}

@book{calsamiglia2005decentralizing,
  title={Decentralizing equality of opportunity and issues concerning the equality of educational opportunity},
  author={Calsamiglia, Caterina},
  year={2005},
  publisher={Yale University}
}

@inproceedings{kusner2017counterfactual,
  title={Counterfactual fairness},
  author={Kusner, Matt J and Loftus, Joshua and Russell, Chris and Silva, Ricardo},
  booktitle={Advances in Neural Information Processing Systems ({NeurIPS} 30)},
  pages={4069--4079},
  year={2017}
}

@article{angwin2016machine,
  title={Machine bias: There’s software used across the country to predict future criminals. and it’s biased against blacks},
  author={Angwin, Julia and Larson, Jeff and Mattu, Surya and Kirchner, Lauren},
  journal={ProPublica, May},
  volume={23},
  year={2016}
}

@article{simoiu2017problem,
  title={The problem of infra-marginality in outcome tests for discrimination},
  author={Simoiu, Camelia and Corbett-Davies, Sam and Goel, Sharad and others},
  journal={The Annals of Applied Statistics},
  volume={11},
  number={3},
  pages={1193--1216},
  year={2017},
  publisher={Institute of Mathematical Statistics}
}

@incollection{collective1977black,
  title={A Black Feminist Statement},
  author={{Combahee River Collective}},
  year={1978},
  booktitle={Capitalist Patriarchy and the Case for Socialist Feminism},
  editor={Zillah Eisenstein},
  publisher={Monthly Review Press, New York}
}

@article{berk2017convex,
  title={A convex framework for fair regression},
  author={Berk, Richard and Heidari, Hoda and Jabbari, Shahin and Joseph, Matthew and Kearns, Michael and Morgenstern, Jamie and Neel, Seth and Roth, Aaron},
  journal={4th Annual Workshop on Fairness, Accountability, and Transparency in Machine Learning},
  year={2017}
}

@misc{truth1851aint,
  title={Ain't {I} A Woman?},
  author={Sojourner Truth},
  note={Speech delivered at Women's Rights Convention, Akron, Ohio},
  year={1851}
}

@article{barocas2016big,
  title={Big data's disparate impact},
  author={Barocas, Solon and Selbst, Andrew D},
  journal={Cal. L. Rev.},
  volume={104},
  pages={671},
  year={2016},
  publisher={HeinOnline}
}

@inproceedings{zhao2017men,
  title={{Men also like shopping: Reducing gender bias amplification using corpus-level constraints}},
  author={Zhao, Jieyu and Wang, Tianlu and Yatskar, Mark and Ordonez, Vicente and Chang, Kai-Wei},
  booktitle={Empirical Methods in Natural Language Processing},
  year={2017}
}

@inproceedings{kearns2018preventing,
	title={{Preventing Fairness Gerrymandering: Auditing and Learning for Subgroup Fairness}},
	author={Kearns, Michael and Neel, Seth and Roth, Aaron and Wu, Zhiwei Steven},
	booktitle={Proceedings of the 35th International Conference on Machine Learning, PMLR 80},
	pages={2569--2577},
	editor = 	 {Dy, Jennifer and Krause, Andreas},
	volume = 	 {80},
	series = 	 {Proceedings of Machine Learning Research},
	address = 	 {Stockholmsmässan, Stockholm Sweden},
	month = 	 {10--15 Jul},
	publisher = 	 {PMLR},
	year={2018}
}

@inproceedings{zafar2017fairness,
	title={{Fairness constraints: Mechanisms for fair classification}},
	author={Zafar, Muhammad Bilal and Valera, Isabel and Rogriguez, Manuel Gomez and Gummadi, Krishna P},
	booktitle={Artificial Intelligence and Statistics},
	pages={962--970},
	year={2017}
}

@inproceedings{friedler2019comparative,
  title={A comparative study of fairness-enhancing interventions in machine learning},
  author={Friedler, Sorelle A and Scheidegger, Carlos and Venkatasubramanian, Suresh and Choudhary, Sonam and Hamilton, Evan P and Roth, Derek},
  booktitle={Proceedings of the conference on fairness, accountability, and transparency},
  pages={329--338},
  year={2019}
}

@article {foulds2020intersectional,
	author = {J. R. Foulds and R. Islam and K. Keya and S. Pan},
	title = {An Intersectional Definition of Fairness},
	journal = {36th IEEE International Conference on Data Engineering (ICDE) (accepted, in press), \textbf{typical acceptance rate 18\%}, ArXiv preprint arXiv:1807.08362 [CS.LG]},
	year = {2020}
}

@article{rosenbaum1983central,
	title={The central role of the propensity score in observational studies for causal effects},
	author={Rosenbaum, Paul R and Rubin, Donald B},
	journal={Biometrika},
	volume={70},
	number={1},
	pages={41--55},
	year={1983},
	publisher={Oxford University Press}
}

@article{rosenbaum1985constructing,
	title={Constructing a control group using multivariate matched sampling methods that incorporate the propensity score},
	author={Rosenbaum, Paul R and Rubin, Donald B},
	journal={The American Statistician},
	volume={39},
	number={1},
	pages={33--38},
	year={1985},
	publisher={Taylor \& Francis Group}
}

@article{eeoc1966guidelines,
	author ={{Equal Employment Opportunity Commission}},
	title = {Guidelines on Employee Selection Procedures},
	location = {Washington, D.C.},
	year = {1978},
	journal = {C.F.R.},
	volume = {29.1607},
}

@article{blodgett2020language,
	title={{Language (Technology) is Power: A Critical Survey of "Bias" in NLP}},
	author={Blodgett, Su Lin and Barocas, Solon and Daum{\'e} III, Hal and Wallach, Hanna},
	journal={arXiv preprint arXiv:2005.14050},
	year={2020}
}

@article{crawford2019ai,
	title={AI now 2019 report},
	author={Crawford, Kate and Dobbe, Roel and Dryer, Theodora and Fried, Genevieve and Green, Ben and Kaziunas, Elizabeth and Kak, Amba and Mathur, Varoon and McElroy, Erin and S{\'a}nchez, Andrea Nill and others},
	journal={New York, NY: AI Now Institute},
	year={2019}
}

@article{foulds2020parity,
  title={Are Parity-Based Notions of {AI} Fairness Desirable?},
  author={Foulds, James R and Pan, Shimei},
  journal={A Quarterly bulletin of the Computer Society of the IEEE Technical Committee on Data Engineering},
  volume={43},
  number={4)},
  year={2020}
}

@inproceedings{zemel2013learning,
  title={Learning fair representations},
  author={Zemel, Rich and Wu, Yu and Swersky, Kevin and Pitassi, Toni and Dwork, Cynthia},
  booktitle={International conference on machine learning},
  pages={325--333},
  year={2013},
  organization={PMLR}
}

@article{ayres2002outcome,
  title={Outcome tests of racial disparities in police practices},
  author={Ayres, Ian},
  journal={Justice research and Policy},
  volume={4},
  number={1-2},
  pages={131--142},
  year={2002},
  publisher={SAGE Publications Sage CA: Los Angeles, CA}
}

@article{kleinberg2016inherent,
  title={Inherent trade-offs in the fair determination of risk scores},
  author={Kleinberg, Jon and Mullainathan, Sendhil and Raghavan, Manish},
  journal={arXiv preprint arXiv:1609.05807},
  year={2016}
}

@inproceedings{pryzant2020automatically,
  title     = {Automatically Neutralizing Subjective Bias in Text},
  author    = {Pryzant, Reid and Martinez, Richard Diehl and Dass, Nathan and Kurohashi, Sadao and Jurafsky, Dan and Yang, Diyi},
  booktitle = {Proceedings of the {AAAI} Conference on Artificial Intelligence},
  volume    = {34},
  number    = {01},
  pages     = {480--489},
  year      = {2020}
}

@article{kelly2023uci,
  title={{The UCI machine learning repository}},
  author={Kelly, Markelle and Longjohn, Rachel and Nottingham, Kolby},
  journal={URL https://archive.ics.uci.edu},
  year={2023}
}

@article{mitchell2021algorithmic,
  title={{Algorithmic fairness: Choices, assumptions, and definitions}},
  author={Mitchell, Shira and Potash, Eric and Barocas, Solon and D'Amour, Alexander and Lum, Kristian},
  journal={Annual review of statistics and its application},
  volume={8},
  number={1},
  pages={141--163},
  year={2021},
  publisher={Annual Reviews}
}

@article{rajkomar2018ensuring,
  title={Ensuring fairness in machine learning to advance health equity},
  author={Rajkomar, Alvin and Hardt, Moritz and Howell, Michael D and Corrado, Greg and Chin, Marshall H},
  journal={Annals of internal medicine},
  volume={169},
  number={12},
  pages={866--872},
  year={2018},
  publisher={American College of Physicians}
}

@article{veinot2018good,
  title={Good intentions are not enough: how informatics interventions can worsen inequality},
  author={Veinot, Tiffany C and Mitchell, Hannah and Ancker, Jessica S},
  journal={Journal of the American Medical Informatics Association},
  volume={25},
  number={8},
  pages={1080--1088},
  year={2018},
  publisher={Oxford University Press}
}

@article{char2018implementing,
  title={Implementing machine learning in health care—addressing ethical challenges},
  author={Char, Danton S and Shah, Nigam H and Magnus, David},
  journal={New England Journal of Medicine},
  volume={378},
  number={11},
  pages={981--983},
  year={2018},
  publisher={Mass Medical Soc}
}

\end{document}